\documentclass[10pt,twocolumn,letterpaper]{article}

\usepackage[pagenumbers]{wacv} 

\usepackage{graphicx}
\usepackage{amsmath}
\usepackage{amssymb}
\usepackage{booktabs}
\usepackage{comment}
\usepackage{float}
\usepackage{lmodern}
\usepackage{adjustbox}
\usepackage{multirow}
\usepackage{pifont}
\usepackage{wrapfig}
\usepackage{bm}
\usepackage{isomath}
\usepackage{array}
\usepackage{makecell}
\usepackage{paralist}

\DeclareMathOperator*{\argmax}{arg\,max}

\def\bo{\mathbf{o}} 

\def\x{\mathbf{x}}

\def\1{\mathbf{1}}

\def\cO{\mathcal{O}} 
\def\cX{\mathcal{X}}

\usepackage[accsupp]{axessibility}  

\usepackage[pagebackref,breaklinks,colorlinks]{hyperref}

\usepackage[capitalize]{cleveref}
\crefname{section}{Sec.}{Secs.}
\Crefname{section}{Section}{Sections}
\Crefname{table}{Table}{Tables}
\crefname{table}{Tab.}{Tabs.}

\def\wacvPaperID{544} 
\def\confName{WACV}
\def\confYear{2025}

\begin{document}

\title{Active Event Alignment for Monocular Distance Estimation}

\author{Nan Cai\\
Technical University of Berlin\\
Germany\\
{\tt\small nan.cai@campus.tu-berlin.de}
\and
Pia Bideau\\
Univ. Grenoble Alpes, Inria\\ 
France \\
{\tt\small pia.bideau@inria.fr}
}
\maketitle

\begin{abstract}
  Event cameras provide a natural and data efficient representation of visual information, motivating novel computational strategies towards extracting visual information. Inspired by the biological vision system, we propose a behavior driven approach for object-wise distance estimation from event camera data. 
  This behavior-driven method mimics how biological systems, like the human eye, stabilize their view based on object distance: distant objects require minimal compensatory rotation to stay in focus, while nearby objects demand greater adjustments to maintain alignment. This adaptive strategy leverages natural stabilization behaviors to estimate relative distances effectively.
  Unlike traditional vision algorithms that estimate depth across the entire image, our approach targets local depth estimation within a specific region of interest. By aligning events within a small region, we estimate the angular velocity required to stabilize the image motion. We demonstrate that, under certain assumptions, the compensatory rotational flow is inversely proportional to the object's distance. The proposed approach achieves new state-of-the-art accuracy in distance estimation - a performance gain of 16\% on EVIMO2. EVIMO2 event sequences comprise complex camera motion and substantial variance in depth of static real world scenes. 
\end{abstract}

\section{Introduction}
\label{sec:intro}

Event cameras mimic certain biological features of the human visual system. Instead of recording RGB frames at a fixed frequency, they record brightness changes as \textit{events} asynchronously and at high temporal resolution. This offers great potential for high-speed automation~\cite{Gehrig2024}, robotics~\cite{hassan2024efficient} and microscopic analysis~\cite{Cabriel2023}, where capturing motion is crucial. Current vision algorithms struggle to efficiently process and interpret asynchronous event data. In this work, we propose to combine the biologically inspired visual acquisition with natural behavioral strategies to enhance the interpretation of event data. Rather than passively analyzing incoming event data at each pixel location, we induce a movement that stabilizes image motion within a targeted region. Similar to eye movements, we introduce rotational adjustments to counterbalance translational camera motion.

\begin{figure}
  \begin{subfigure}{.63\linewidth}
    \centering
    \includegraphics[width=6cm]{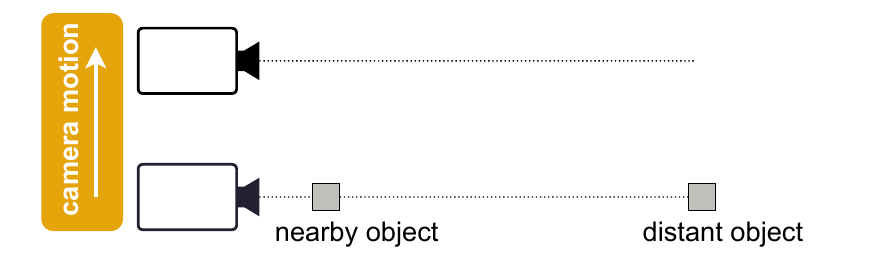}
    \caption{\scriptsize Camera motion \\ \textcolor{white}{blabla blabla blabla blabla}}
  \end{subfigure}
  \hspace{1em}
  \begin{subfigure}{.3\linewidth}
    \centering
    \includegraphics[width = \linewidth, height=1.85cm]{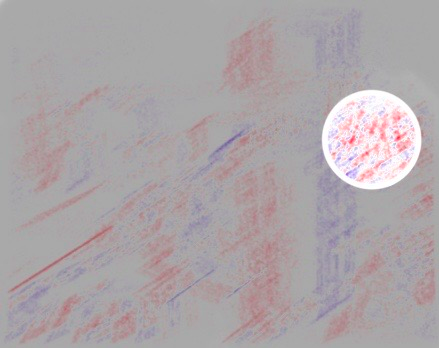}
    \caption{\scriptsize Observed events \textcolor{white}{blabla}}
  \end{subfigure}
  \begin{subfigure}{.63\linewidth}
    \centering
    \includegraphics[width=6cm]{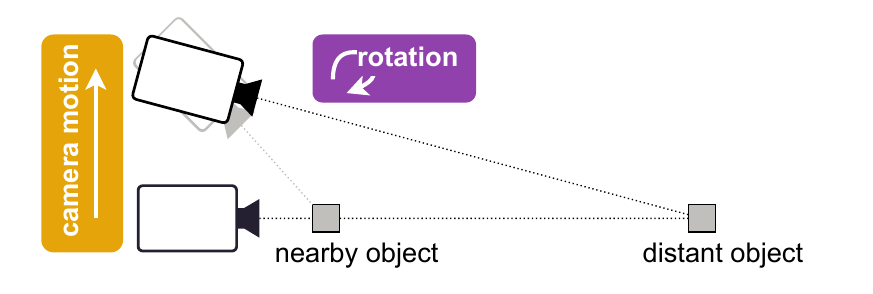}
    \caption{\scriptsize Camera motion + rotation \\ \textcolor{white}{blabla blabla blabla blabla}}
  \end{subfigure}
  \hspace{1em}
  \begin{subfigure}{.3\linewidth}
    \centering
    \includegraphics[width = \linewidth, height=1.85cm]{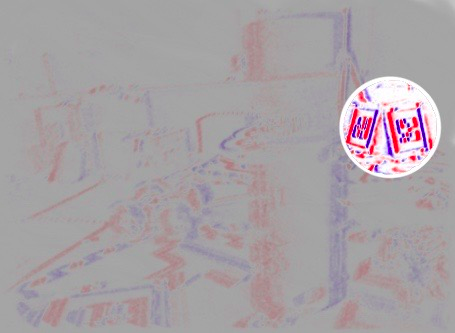}
    \caption{\scriptsize Aligned events \\ \textcolor{white}{blabla}}
  \end{subfigure}
  \vspace{-2em}
  \caption{Akin to gaze stabilization, active event alignment stabilizes a local image region by applying a rotation that counteracts the camera's motion. This rotation leads to locally well aligned events as pictured in (d). The relative distance between two objects can then be inferred by comparing their compensatory rotations. }
  \label{images:gaze-stabilization}
\end{figure}

\textit{What are the benefits of stabilization?}
From an ecological perspective, stabilization reduces the number of brightness changes detected by the eye's photoreceptors. As photoreceptors (``light sensors'' of the human eye) are slow and capture only a small part of the scene briefly, fast body movements can significantly degrade vision unless counteracted by eye movements~\cite{land2012animal}. By doing so stabilization conveys relevant scene-specific information, such as relative distances between objects.
This paper proposes a novel method for distance estimation by \textit{active event alignment}. To stabilize incoming visual information within a specific area of the camera's sensor, we introduce rotational motion that compensates for the observer's movement. Under certain assumptions that are discussed in this paper, the compensatory rotational flow is inversely proportional to the object’s distance. An example is shown in ~\cref{images:gaze-stabilization}. Intuitively speaking, the farther away an object the less compensatory motion is needed for stabilization on the camera sensor. Thus the amount of compensatory motion needed to stabilize an object bears strong cues for 3D object localization.

We propose a novel alignment strategy, that unlike prior work \textit{does not} aim at reconstructing the camera's motion. Instead, it determines the rotational velocity that leads to best event alignment across a set of small image regions, from which the object's relative distances are inferred.
Without computing explicit correspondences between events and without knowledge about the camera's pose, we measure the degree to which a set of events are jointly aligned. The quality of the joint event alignment is assessed by calculating pixelwise entropy, leveraging redundant information to achieve a more effective representation~\cite{field1994goal,barlow1999finding,barlow1989unsupervised}. 

\noindent{\bf Contributions.} We propose a novel approach for relative distance estimation through object-wise event alignment. Our two step optimization strategy for event alignment robustly estimates a compensatory rotation that is directionally consistent but varies in velocity across the image. Relative depth is then computed by comparing the compensatory rotations of the object and the reference region. Notably, our method relies only on relative camera motion, not absolute camera poses.
The approach is supported by new state-of-the-art results on a \textit{de facto} benchmark - EVIMO2 for relative object-wise depth estimation. 
The approach aligns with object-centered visual representations, which are beneficial for various tasks~\cite{roth2023objects, ballard1990animate, mishra2009active}.
\newline
\textbf{Code will be made publicly available.}

\section{Related Work}
Previous work on distance estimation using event cameras does not focus on behavioral strategies to extract information about the scene's structure - like relative distances among objects. We review general active perception methods (not necessarily relying on event data) and common depth estimation approaches using single or stereo event cameras.

\paragraph{Active perception.} Active perception assumes the observer to be active - ``The purpose of the activity is to manipulate the constraints underlying the observed phenomena in order to improve the quality of the perceptual results''~\cite{aloimonos1988active}. 
The constraints of fixation and tracking have been studied in~\cite{fermuller1992tracking,fermuller1993role}.
Following similar principles, Burner et al.~\cite{burner2023ttcdist}, for the first time, provides a closed-form solution for the estimation of the distance from a sliding-window of time-to-contact and inertial measurements (IMU). Their closed-form solution relies on non-constant acceleration during the time interval of computation. In contrast, our approach computes nearly instantaneous depth estimates over small time intervals, requiring only translational motion and no acceleration constraints.
Battaje et al.~\cite{battaje2022one} address distance estimation with RGB cameras by pre-selecting a region and introducing an action to enhance visual processing. Similarly, we exploit the dynamic nature of event-based vision sensors for distance estimation through active vision.

\paragraph{Event based distance estimation} Distance estimation has been a long-standing challenge for event cameras. We review distance estimation methods using Multi-View Stereo, Deep Learning, Neuromorphic Processing, and Active Vision.

\textit{Multi-View Stereo.} Depth estimation with stereo event cameras relies on the epipolar constraint and the assumption that events occur simultaneously on both cameras. Zhou et al. propose an event based SLAM approach ~\cite{zhou2021event_stereoVO}. In particular, they reformulate temporal coincidence of events using the compact representation of space-time provided by time surfaces~\cite{HOTS}. Inspired by~\cite{rebecq2018emvs}, Ghosh et al.~\cite{Ghosh_2022} circumvent the challenge of accurate event association by leveraging the sparsity of events and by exploiting the continuity of camera viewpoints. Using Space Sweeping, it builds ray density Disparity Space Images (DSIs) from each camera data and fuses them into one DSI~\cite{DSI}. A small set of work exploits Multi-View with only a single event camera. Here, event correspondences across time are established via alignment~\cite{gallego2018unifying,gallego2017accurate} and under the limiting assumptions that the camera pose is known and the scene is static. Rebecq et al. ~\cite{rebecq2018emvs} similarly relies on knowledge about the camera's absolute motion trajectory. A depth value is associated to each event resulting in a semi-dense depth map.

\textit{Deep Learning.} Hidalgo-Carrio et al.~\cite{hidalgo2020learning} introduced the first supervised method to learn dense monocular depth from event data. Various self-supervised approaches have been introduced to learn dense depth maps. Zhu et al.\cite{zhu2023self} exploit cross-modal consistency between frames and aligned events, while Zhu et al.\cite{zhu2019unsupervised} predict egomotion and depth by minimizing motion blur when events are projected onto the image plane. The supervision signal in the latter comes from motion compensation through event alignment.
A different line of work by Rudnev et al.~\cite{rudnev2023eventnerf} and Hwang et al.~\cite{hwang2023ev} investigates how NeRFs could be reconstructed from event data. 



\begin{table*}
\centering
\small
\begin{adjustbox}{max width=0.99\linewidth}
\setlength{\tabcolsep}{3pt}
\begin{tabular}{lc ccc ccc ccc}
\toprule 
Method & Algorithmic & Monocular & Input Data & Absolute & Relative & Data Structure of & Freq.& \\
& Approach & Camera &  & Cam. Pose & Cam. Motion & Estimated Depth & [in Hz] & \\
\midrule
E2Depth~\cite{hidalgo2020learning} & Deep Learning & \ding{51} & Events & \ding{55} & \ding{55} & Pixel-wise Depth & 20  & \\ 
EMVS~\cite{rebecq2018emvs} & Multi-View Stereo & \ding{51} & Events & \ding{51} & \ding{51} & Event-wise Depth & 1 & \\ 
Ours & Active Vision & \ding{51} & Events & \ding{55} & \ding{51} & Object-wise Depth & 20 &  \\
\bottomrule
\end{tabular}
\end{adjustbox}
\vspace{0.5ex}
\caption{Literature review on event-based depth estimation approaches. We characterize each algorithm that was used for our evaluation according to its properties: algorithmic approach, monocular vs. stereo camera setup, input data, required additional sensor information such as absolute camera pose or relative camera motion, the format of the algorithm's depth estimates as well as its frequency it was evaluated on. 
}
\label{table:literature_review}
\end{table*}

\section{Distance Estimation via Region-wise Event Alignment}
We present our approach for distance estimation from event data. We start with revising key aspects of the probabilistic model \textit{``the Spatio-Temporal Poisson Point Process for event alignment''} from Gu et al.~\cite{Gu_2021} and propose small - yet effective changes that increase robustness with respect to variations in speed and scene structure. This framework is applied to object-wise event alignment, detailed in Section~\ref{subsubsec:object-wise-alignment}. Different from classical event alignment approaches~\cite{gallego2017accurate, Nunes20eccv,gallego2019focus,stoffregen2019event}, our goal is not to recover the camera's motion. Instead, we introduce a novel compensatory rotational motion that stabilizes the motion within a region of interest. While other approaches for event alignment could be adopted for local alignment, we build on~\cite{Gu_2021}, as this algorithm is currently best performing and methodological consistent with ideas introduced in this work. More specifically, events are aligned by minimizing the pixel-wise entropy resulting in reduced brightness changes per pixel. This implements an ecological principle that aims at reducing perceived brightness changes on the human eye's retina though gaze stabilization~\cite{field1994goal,barlow1999finding,land2012animal}.
In Section~\ref{subsubsec:distance} we introduce a novel formulation, that relates a rotational velocity estimated by local event alignment to object distance. Intuitively speaking, the smaller the rotational motion is to align a specific object-region, the farther the object-region away (see Figure~\ref{images:gaze-stabilization}). This approach for the first time allows distance estimation, without computing explicit event-to-event correspondences and without knowledge about the absolute camera's pose.

\subsection{Event Alignment}
Event alignment describes the process of finding a transformation $\mathcal{T}_{\omega}$ that maps events triggered by the same world point to the same pixel location of the camera sensor.
We define
\begin{equation}
\mathcal{O}=[\bo_1,\bo_2,\ldots,\bo_N],
\end{equation}
to be a set of $N=|\cO|$ events, where each element $\bo_i=(o^x_i, o^t_i, o^p_i)$ comprises a pixel location $o^x_i$, timestamp $o^t_i$ at which the event occurs and a polarity $o_p$ determining the direction in brightness change.
Most commonly, event alignment is formulated as an optimization problem over rotational camera motion parameters $\omega$ and translational parameters $v$,
\begin{equation}
\hat{\omega}, \hat{v}\,=\,\argmax_{\omega, v} p\big(\mathcal{T}_{\omega, v}(\cO)\big).
\end{equation}
While there have been proposed a number of different loss functions for solving this optimization problem, we implement the loss by Gu et al. In their method, Gu et al~\cite{Gu_2021}. model an aligned event stream at a particular pixel location as a \textit{Poisson Point Process}. Based on this model, a maximum likelihood approach is developed to register events that are initially unaligned. We find the transformations $\mathcal{T}$ of the observed events $\cO$ that make them as
likely as possible under the model:
\begin{equation}
\label{eq:uv_alignment}
p\big(\mathcal{T}_{\omega, v}(\cO)\big) = \prod_{\x \in \cX} \mathcal{NB}\big(k_\x(\mathcal{T}_{\omega, v})\big).
\end{equation}
Here, $k_{\x}$ denotes the number of events occurring at a location $\x$ and is modeled with a negative binomial distribution $\mathcal{NB}(\cdot)$\footnote{The negative binomial distribution arises as a mixture of Poisson distributions where the Poisson rate parameter itself follows a gamma distribution. In other words, we can view the negative binomial as a $\text{Poisson}(\lambda)$ distribution, where $\lambda$ is itself a random variable, distributed as a gamma distribution.}. 
In Gu et al., $\cO$ is defined as a set of $N$ discrete events. In contrast, we redefine $\cO$ as a set of events occurring within a fixed time interval $\Delta T$. This modification aligns with the classical Poisson Point Process definition, better capturing event dynamics. Its effectiveness is demonstrated in \cref{sec:experiment}.

\begin{figure*}[t]
    \centering
    \includegraphics[width=\linewidth]{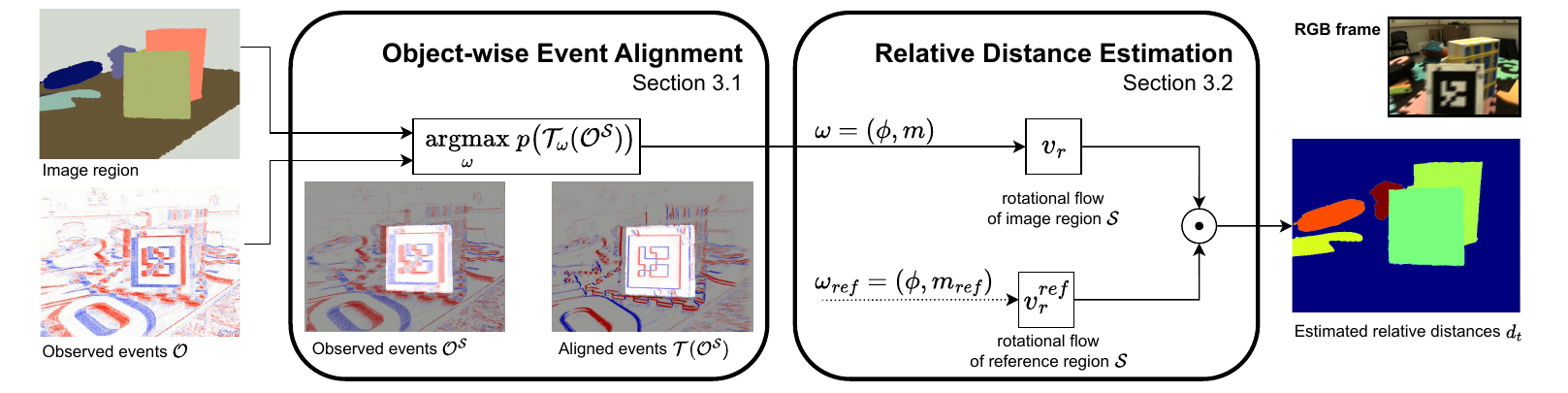}
    \caption{Overview of our algorithm. We estimate relative object-wise-distance from active event alignment. Given a set of events, we process alignment in a object-wise fashion. Object regions maybe determined by a provided object segmentation mask or a default segmentation mask (e.g honeycomb) without semantic information. The obtained relation between different angular velocities within the image plane determine the respective object-wise relative depth.}
    \label{images:pipeline}
\end{figure*}

\subsubsection{Object-wise Event Alignment}
\label{subsubsec:object-wise-alignment}

Events arise by the relative motion of the camera and the scene. The events
recorded at a single sensor location seldom correspond to the same world point. Classical event alignment algorithms aim to register events triggered by the same world point through estimating the camera’s motion. Opposed to classical event alignment, we do not aim at recovering the camera's motion through event alignment, instead we wish to find a rotational velocity $w$ that locally leads to accurate event alignment. Although the estimated rotational velocity does not relate to the true camera motion, that triggered the observed events, we will show in \cref{subsubsec:distance} that this virtual rotation carries valuable information about relative distances across different image regions.

The virtual rotational velocity $w$ is estimated via region-wise event alignment, that is:
\begin{equation}
\label{eq:alignment_fixation}
\hat{\omega}\,=\,\argmax_{\omega} p\big(\mathcal{T}_{\omega}(\cO^\mathcal{S})\big),
\end{equation}
$\mathcal{S}$ denotes the local image region. This optimization leverages the concept that, for rigid camera motion and fronto-parallel planar scene regions, a camera translation\footnote{In case of additional camera rotation, IMU information is used to remove the rotational motion part.} can be well approximated by a virtual rotation, and vice versa~\cite{mithokinIROS18}.
Limiting the possible transformation $\mathcal{T}$ to rotation only motions, comes with the benefit of a significantly reduced amount of parameters to be estimated, on the other hand if the optimization is confined to a small image region only, this may lead in unstable rotation estimates, that do not serve as a surrogate of the true camera translation.
To this end, we propose a strategy to perform object-wise alignment in two steps: (1) we determine a global \textit{velocity direction} that performs alignment across all objects present in the scene and (2) we determine the \textit{velocity's magnitude} that aligns the events of a specific object. This strategy takes the best of both worlds - it determines the global velocity direction over a large image region while assessing the velocity's magnitude for specific (potentially small) image regions, thus maintaining small parameters to be estimated and achieving robust rotation estimates.

\paragraph{Estimation of the global velocity's direction.} According to physical rules of perspective geometry, a local motion direction is independent upon the scene depth in case of pure camera translation~\cite{Bideau2024}. Thus for planar scenes the motion direction can be well approximated via a rotational camera motion~\cite{mithokinIROS18}. In scenes with significant depth variations, objects that are closer tend to exhibit larger motion magnitudes, while those that are farther away show smaller motion magnitudes on the image plane. One option to deal with unknown depth is to estimate pixel-wise depth values through maximum likelihood estimation alongside $\omega$. This would drastically increase the number of parameters to be estimated. Instead, we adopt a partially Bayesian approach by maximizing the marginal likelihood of $k_{\bf x}$ under the velocity's magnitude $m$. The angular velocity is expressed in polar form $\omega = (m, \phi)$. Consequently, the direction $\phi$ can be determined by integrating over the unknown magnitude $m$:
\begin{align}
\label{eq:marginalization}
p\big(\mathcal{T}_{\omega}(\cO^{\bf x})\big) &= \int_{m} p(k_\x (m,\phi)| m) \cdot p(m)\,dm\\
&= \int_{m} \mathcal{NB}_{r,q}\big(k_\x(m,\phi)\big) \cdot \mathcal{U}_{[0,m_{\text{max}}]}\,dm\\
&= \int_{m} \mathcal{NB}_{r,q}\big(k_\x(m,\phi)\big)\,dm
\end{align}
An estimate of the velocity's direction $\phi$ is obtained by maximizing the probability of aligned events:
\begin{equation}
\label{eq:obj_func}
\hat{\phi}\,=\,\argmax_{\phi} p\big(\mathcal{T}_{\omega}(\cO)\big)  =\,\argmax_{\phi} \prod_{\x \in \cX} p\big(\mathcal{T}_{\omega}(\cO^{\bf x})\big).
\end{equation}
The maximization is applied across all pixel locations $\bf x \in \cX$ on the image plane. Afterwards, depth discontinuities are handled by calculating the velocity's magnitude (speed) for each object. 

\paragraph{Estimation of the velocity's magnitude.} Given the estimate of the velocity's direction $\hat{\phi}$, the object's magnitude is estimated in a similar fashion:
\begin{equation}
\label{eq:mag_obj_func}
\hat{m}\,=\,\argmax_{m} p\big(\mathcal{T}_{\omega}(\cO^{\mathcal{S}})\big).
\end{equation}
The maximization is applied across all pixel locations $\bf x \in \mathcal{S}$ on within an object region.
The advantage of sequential alignment is two-fold. First, obtaining an estimate of the velocity's direction across the image introduces an additional geometric constraint. Second, this approach significantly reduces the number of parameters that need to be optimized.

\subsection{Relative Distance Estimation}
\label{subsubsec:distance}

The estimated rotational velocity $\omega$ counteracts the camera's translational movement, stabilizing the image within a specific region $\mathcal{S}$. If the object region is distant, stabilization requires minimal compensatory rotational movement. In contrast, if the object region is nearby, greater compensatory rotation is needed for accurate alignment.
Mathematically, we can express this stabilization in terms of optical flow\footnote{Optical flow $\vectorsym{v}$ represents the camera's projected motion onto the image plane. Given the camera's motion, Horn et al.~\cite{Horn1981} provide equations to determine the flow at a specific pixel location.}. To formalize this, we make the following model assumptions:
\begin{compactenum}[i)]
    \item Fronto-parallel planar object regions
    \item Zero translational motion along depth axis
\end{compactenum}
\begin{align}
\label{eq:gaze_stabilization}
\frac{1}{z}\vectorsym{v_t} &\overset{\text{i),ii)}}{=} -\vectorsym{v_r}\\
z &= -\vectorsym{v_r^{+}}\vectorsym{v_t}.
\end{align}
Numbers over equality signs give the assumption that is invoked. Object-wise alignment results in zero local flow. More specifically, the rotational flow $\vectorsym{v_r}$ compensates the translational flow $\vectorsym{v_t}$ leading to zero local flow, which is indicated by Eq~\eqref{eq:gaze_stabilization}. This concept enables a behavior-driven approach to estimating relative distances from event camera data. 
By relating the two rotational flow vectors, the relative distance $d$ between two objects can be inferred:
\begin{equation}
    d = \frac{z}{z^{\text{ref}}}
    \overset{\text{ii)}}{=} \vectorsym{v^{\text{ref}}_r}\vectorsym{v_r^{+}},
\end{equation}
where $\vectorsym{v_r^{+}}$ is the pseudo-inverse of the estimated rotational flow vector, and $\vectorsym{v_r^{ref}}$ is the rotational flow vector estimated for the reference object. The reference object is defined as the largest region in the scene. Assumption ii) renders the translational flow invariant to the object's position. Consequently, the division can be simplified by canceling out translational flow.

\subsubsection{Recursive Bayesian Filtering}

To maintain temporal consistency despite varying camera motions and occlusions, we employ Bayesian filtering. The recursive nature of Bayes filters enables continuous processing of measurements as they arrive~\cite{thrun2002probabilistic}. A belief over the relative distance is propagated over time in two steps: 1) prediction and 2) update using new observation $o_t$. 
\begin{align}
p\big(d_t| o_{t-1}\big) &= \int p\big(d_t| d_{t-1}\big) \label{eq:kalman_predict}
p\big(d_{t-1} | o_{t-1}\big)\,d d_{t-1}\\
p\big(d_t | o_t\big) &= \eta p\big(o_t|d_t\big) p\big(d_t| o_{t-1}\big)\label{eq:kalman_update}
\end{align}
This notation highlights the recursive nature: the posterior from the previous step $p(d_{t-1}|o_{t-1})$ is used to predict the current posterior at time $t$, which is then updated in Eq.~\eqref{eq:kalman_update} using the latest observation.
The estimated relative distance is treated as a Gaussian distribution, and a 1D Kalman filter is used to track the distance and its variance over time.
The variance is modeled as $\sigma^2=\frac{1}{|\vectorsym{v_r}|^2}$, depending on the magnitude of the compensatory rotational motion $\vectorsym{v_r}$.
Further implementation details are in suppl. material.

\begin{figure}[t]
    \centering
    \includegraphics[width=\linewidth]{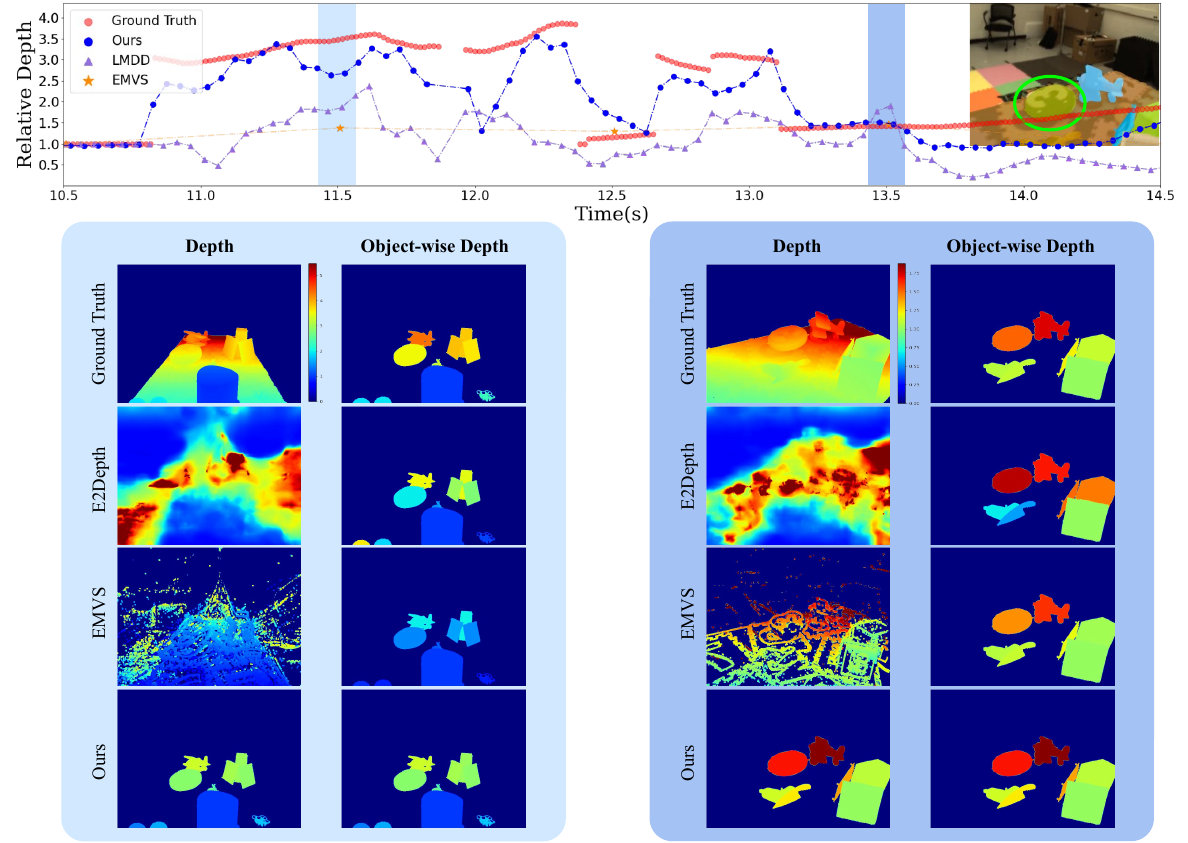}
    \caption{\textbf{Qualitative results} of object-wise relative depth estimation over time. Top: The line plot shows the relative depth estimation of the object with ID 24 (highlighted with a green circle) of the event sequence \textit{scene\_03\_00\_000000}. 
    }
    \label{images:depth_comp1}
\end{figure}

\section{Experiments}
\label{sec:experiment}

We present evaluation results of our monocular distance estimation method on the EVIMO2 dataset~\cite{EVIMO2}, and improvements in camera velocity estimation on the DAVIS 240C dataset~\cite{Mueggler17ijrr}. We describe the datasets, experimental setup, evaluation metrics, and discuss our findings on event-based distance estimation via object-wise alignment.

\noindent{\bf EVIMO2}~\cite{EVIMO2} is a widely used dataset for evaluating event-based algorithms on depth estimation and object segmentation.
The sequences feature the objects on a board, under different lighting conditions.
The recording setup of EVIMO2 includes three VGA-resolution event-based cameras, namely, a Samsung DVS Gen3 in the middle, two Prophesee cameras on the left and right side.
Additionally, a Flea3 frame-based RGB camera is also mounted.
In this work, we use the data from the Samsung DVS Gen3 event camera and from the IMU embedded in the Prophesee camera.
EVIMO2 has various types of sequences for different event-based vision tasks under different conditions.
In our experiments, we adopt the ones for structure from motion, where the GT camera poses (not used), object segmentation and scene depth are available.
To showcase event cameras' capabilities in varying lighting, we use ten sequences recorded in normal light and two out of five under low-light conditions, totaling 147s and 82s, respectively. Only two low-light sequences include IMU sensor data and object masks.

\noindent\textbf{The DAVIS 240C dataset~\cite{mueggler2017event}} includes four 60-second sequences that showcase rotational motion: shapes, poster, boxes, and dynamic. Initially, the camera rotates around each axis at increasing speeds, followed by free rotation in 3 degrees of freedom (3-DOF). 
The shapes, poster, boxes sequence capture static indoor scenes with varying levels of texture complexity, generating approximately 20 to 200 million events. The dynamic sequence, features a dynamic scene and produces around 70 million events.

{\bf Evaluation metrics.}
Following the commonly used evaluation protocols of event-based and frame-based depth estimation in literature~\cite{shi2023even,jiang2018self,godard2019digging}, four error metrics are employed: RMSE (linear), RMSE (log), squared relative distance (SRD) and absolute relative distance. The explicit mathematical formulation of each utilized measurement can be found in~\cite{2014depth}. Accuracy is measured using three thresholds that determine the percentage of estimates with relative accuracy $\delta$ below each threshold. If $\delta = 1$, the estimate matches the ground truth perfectly. The higher the threshold the more values fall into the bin of relative accuracy lower than the provided threshold.



{\bf Implementation details.} We use a fixed $\Delta T$ of 0.05 seconds (20Hz) for all our experiments. All model parameters required for alignment remain consistent with those from Gu et al.~\cite{Gu_2021}, but we reduce the maximum iterations from 250 to 50 to speed up event alignment.
The process noise's standard deviation $\sigma$ of the Kalman filter is set to 0.1. 

\subsection{Results}
\label{sec:results}

\begin{table*}[t]
\centering
\small
\begin{adjustbox}{max width=0.99\linewidth}
\setlength{\tabcolsep}{3pt}
\begin{tabular}{llccccccccc}
\toprule 
\multicolumn{1}{l}{} &
\multicolumn{2}{l}{}    &
\multicolumn{4}{c}{Error $\downarrow$}   &
\multicolumn{3}{c}{Accuracy $\uparrow$}   &\\ 
\cmidrule(lr){4-7}
\cmidrule(lr){8-10}
 & Method & Freq. [Hz] & RMSE (linear) & RMSE (log) & ARD & SRD & $\delta < 1.25$ [\%] & $\delta < 1.25^2$ [\%] & $\delta < 1.25^3$ [\%]\\
\midrule 
sfm & E2Depth~\cite{hidalgo2020learning} & 20 & 1.042 & 0.667  & 0.428 & 0.606 & 35.791 & 54.284 & 71.035\\ 
(10 Seq.)& EMVS~\cite{rebecq2018emvs} & 1 & 0.871& 0.621 & 0.401 & 0.496 &40.775 & 59.460&70.369\\ 
& Ours & 20 & \bf{0.725} & \bf{0.448} & \bf{0.273} & \bf{0.293} & \bf{56.308} & \bf{80.786} & \bf{90.540}
 \\[0.15cm]
sfm & E2Depth~\cite{hidalgo2020learning} & 20 &  1.034 & \bf{0.567} & 0.414 & 0.609 & \bf{47.276} & \bf{64.790} & \bf{79.701} \\
{\footnotesize (low light)} & EMVS~\cite{rebecq2018emvs} & 1 & 0.928 & 0.630 & \bf{0.404} & 0.424 & 39.860 & 61.111 & 71.018 \\
(2 Seq.)& Ours & 20 & \bf{0.883} & 0.930 & 0.414 & \bf{0.420} & 41.077 & 56.728 & 68.168 \\
\bottomrule
\end{tabular}
\end{adjustbox}
\vspace{0.5ex}
\caption{Relative object-wise depth estimation of static scenes with multiple objects. Accuracy comparison on event sequences from EVIMO2~\cite{EVIMO2} - sfm and sfm (low light). 
}
\label{table:main-results}
\end{table*}

We evaluate our approach using the publicly available EVIMO2 dataset. We present the results and an ablation study to demonstrate our method for distance estimation from event camera data. Our approach utilizes event data, rotational velocity from the IMU to compensate for camera rotation, and object segmentation masks to identify regions of local alignment. Ablation studies show that while object segmentation masks can be beneficial, the proposed algorithm effectively estimates scene depth even with arbitrary masks.

\def\figWidth{0.2\linewidth}
\begin{figure}
	\centering
    {\small
    \setlength{\tabcolsep}{0.5pt}
    
	\begin{tabular}{
	>{\centering\arraybackslash}m{0.195\linewidth} 
	>{\centering\arraybackslash}m{\figWidth}
	>{\centering\arraybackslash}m{\figWidth}
 	>{\centering\arraybackslash}m{\figWidth} 
	>{\centering\arraybackslash}m{\figWidth} 
	>{\centering\arraybackslash}m{0.2cm} 
        }
        \includegraphics[width=\linewidth]{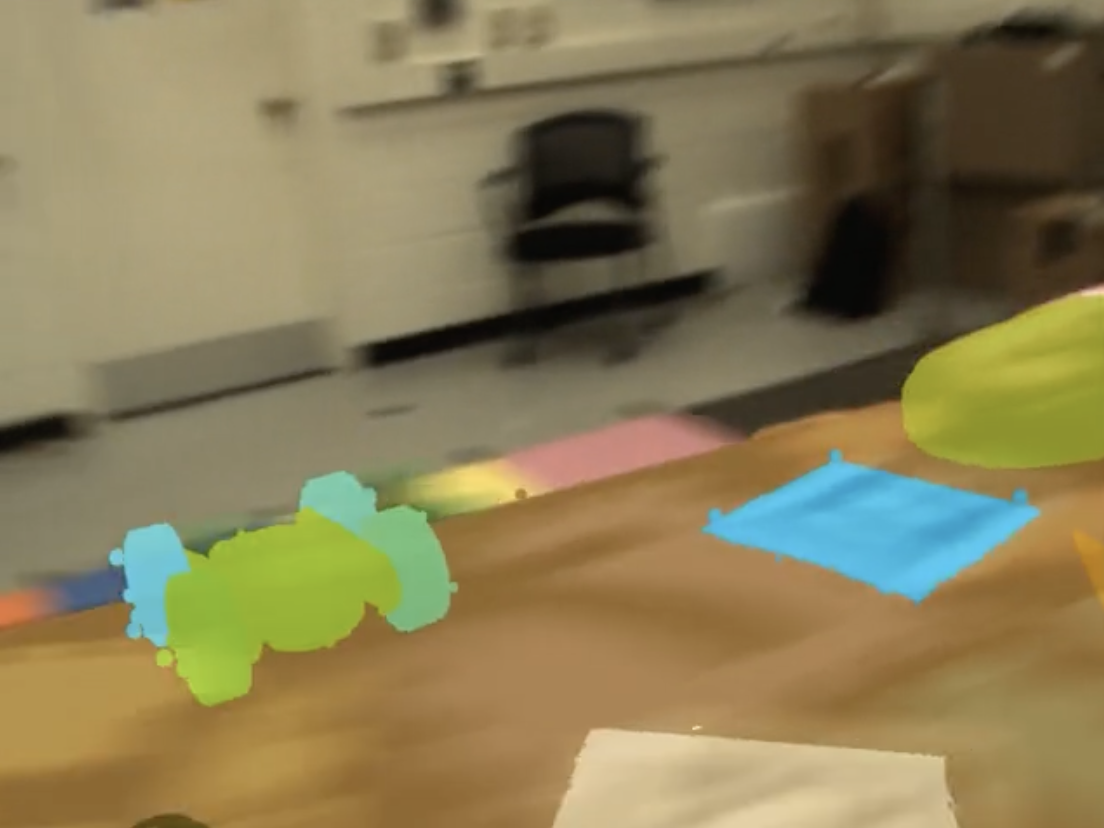}
            &\includegraphics[width=\linewidth]{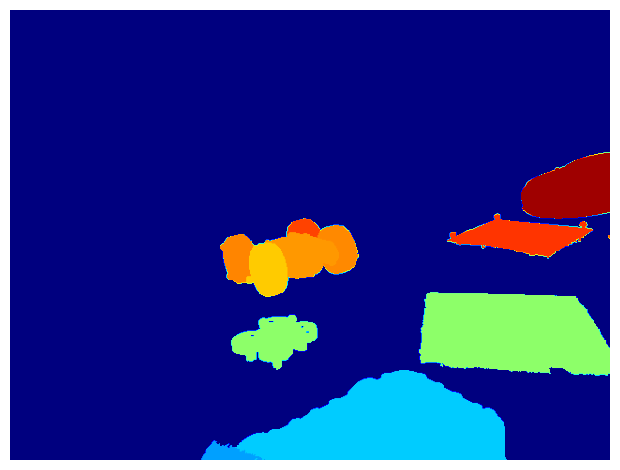}
            &\includegraphics[width=\linewidth]{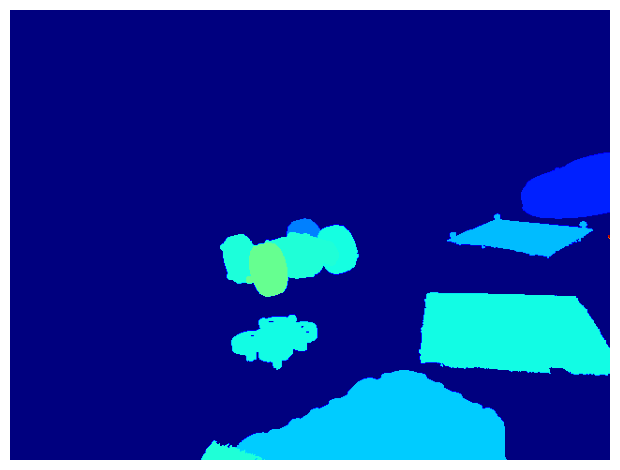}
            &\includegraphics[width=\linewidth]{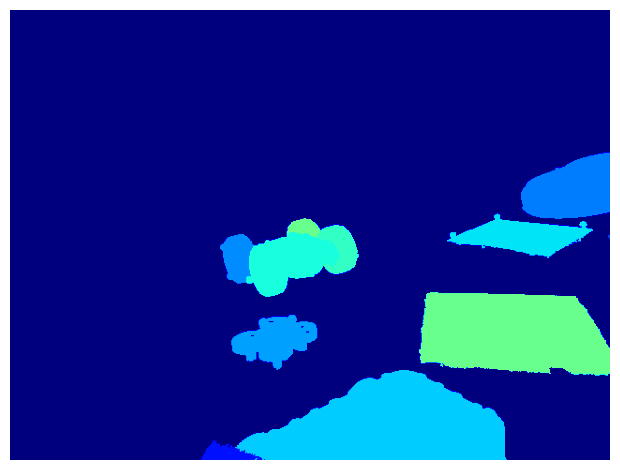}
            &\includegraphics[width=\linewidth]{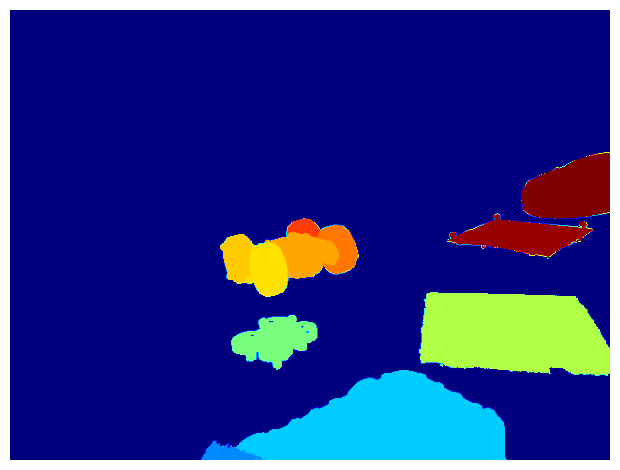}
            &\includegraphics[width=\linewidth]{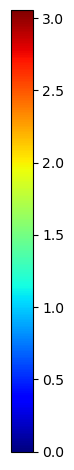}
		\\
        \includegraphics[width=\linewidth]{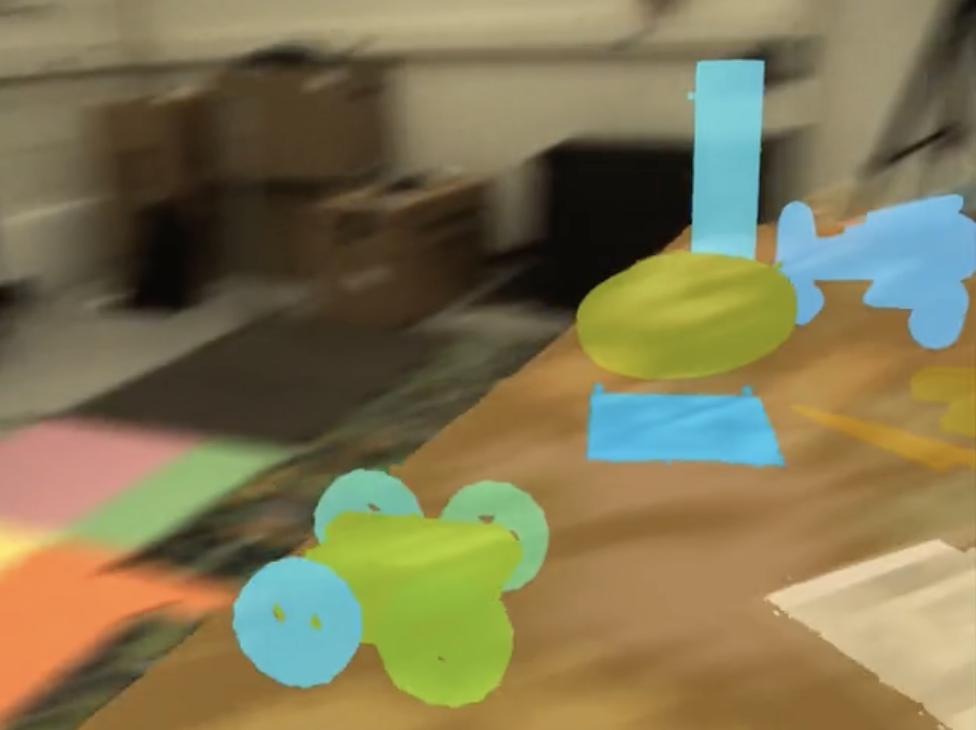}
            &\includegraphics[width=\linewidth]{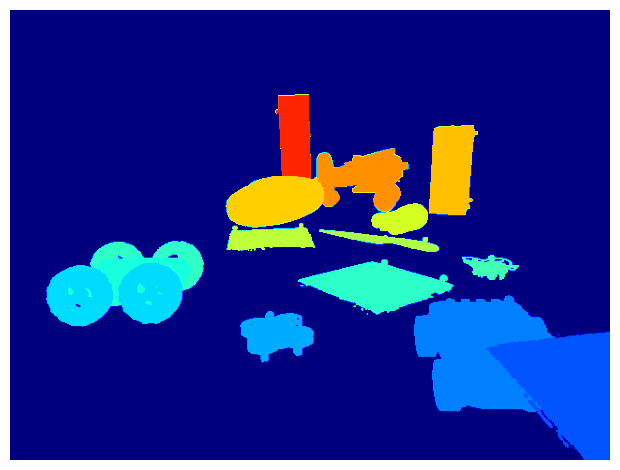}
            &\includegraphics[width=\linewidth]{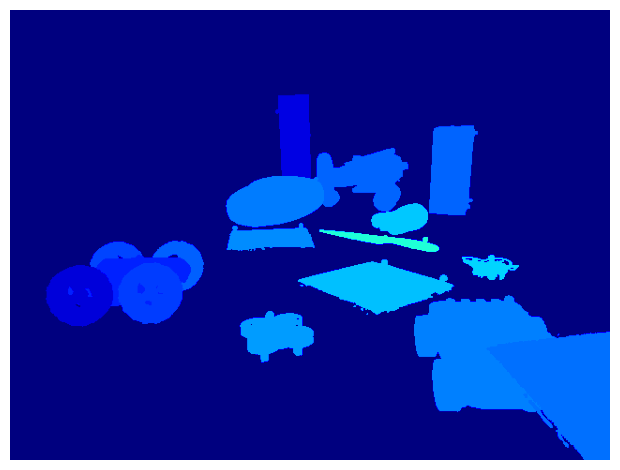}
            &\includegraphics[width=\linewidth]{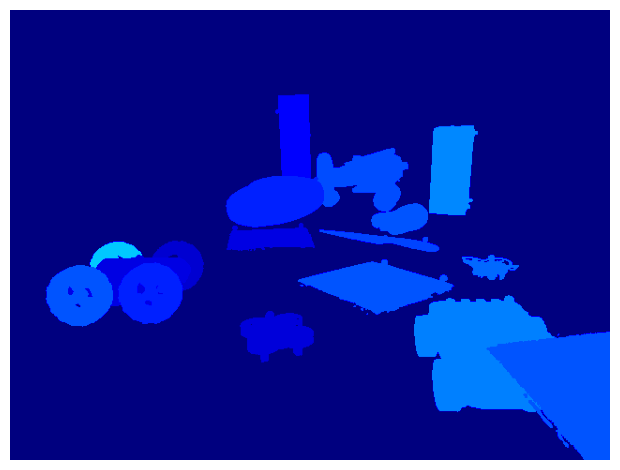}
            &\includegraphics[width=\linewidth]{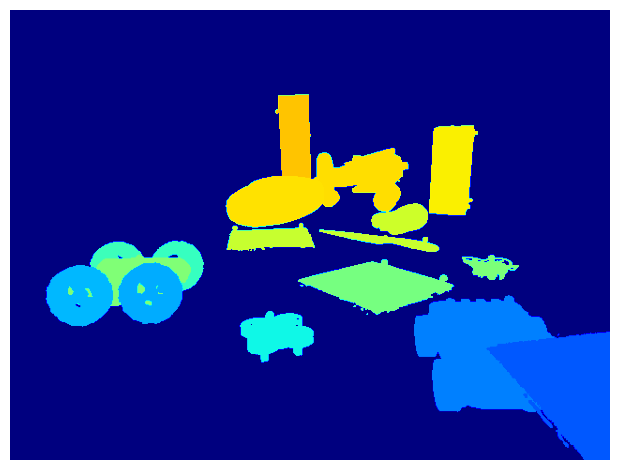}
            &\includegraphics[width=\linewidth]{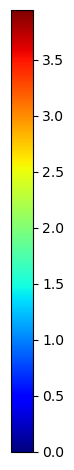}
		\\
        \includegraphics[width=\linewidth]{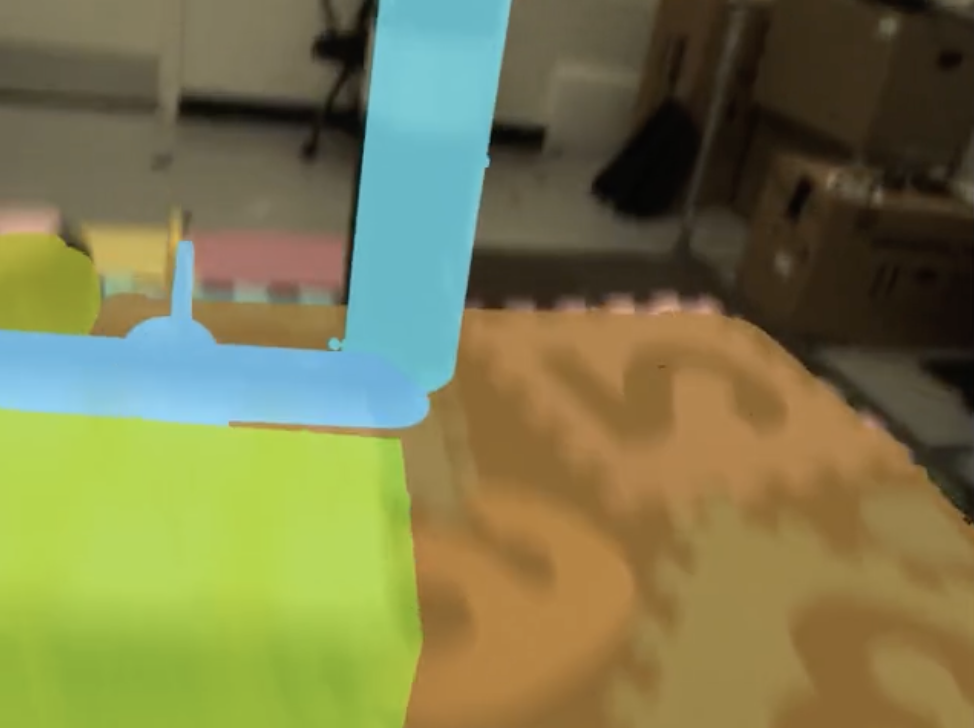}
            &\includegraphics[width=\linewidth]{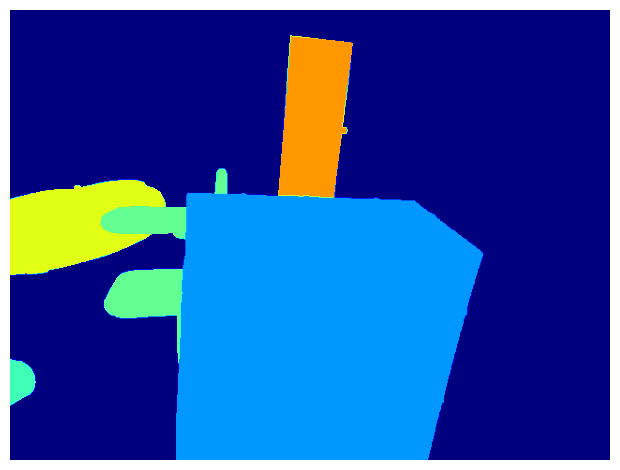}
            &\includegraphics[width=\linewidth]{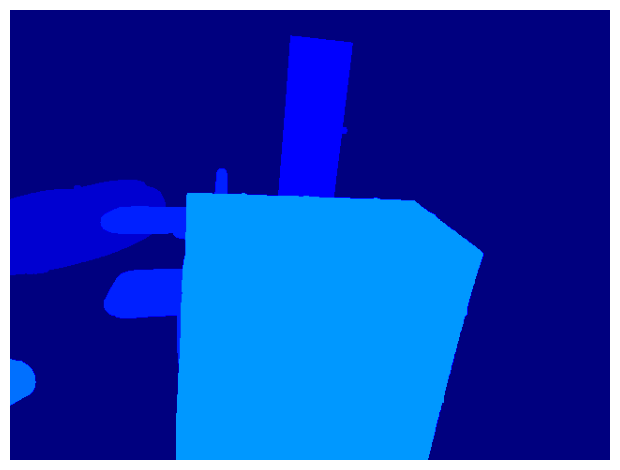}
            &\includegraphics[width=\linewidth]{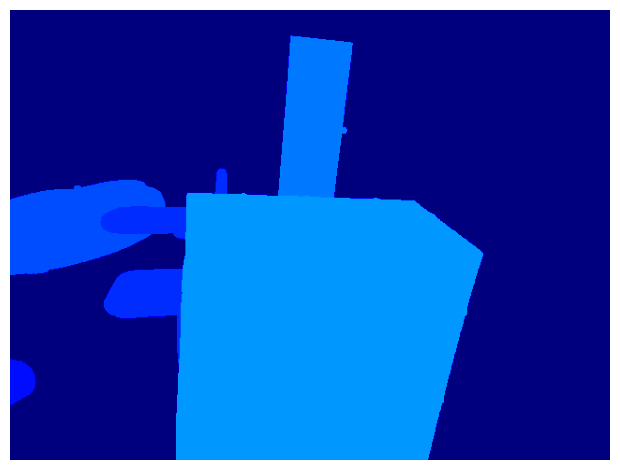}
            &\includegraphics[width=\linewidth]{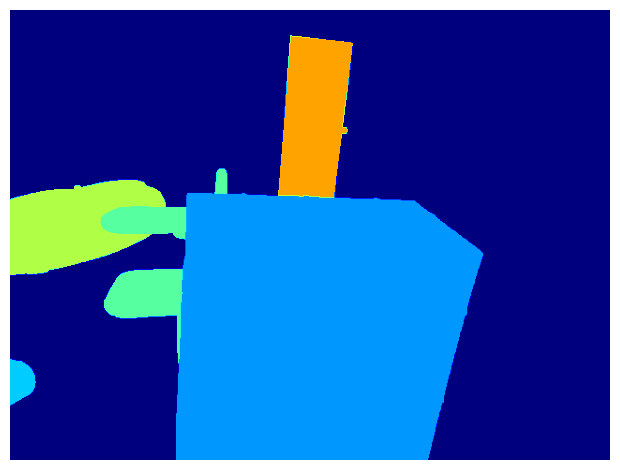}
            &\includegraphics[width=\linewidth]{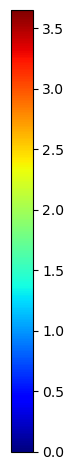}
		\\
        \includegraphics[width=\linewidth]{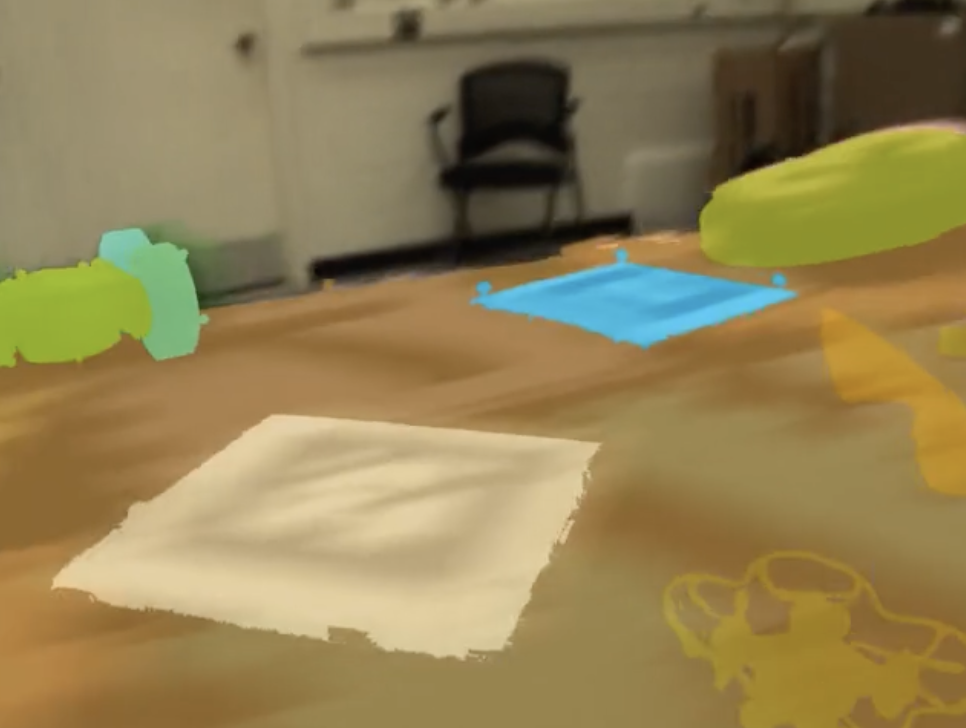}
            &\includegraphics[width=\linewidth]{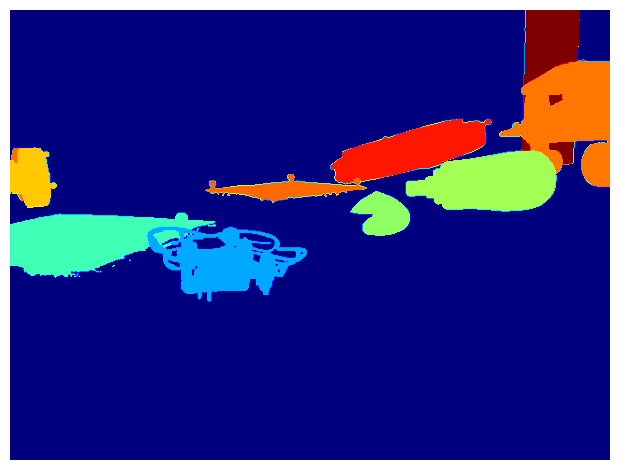}
            &\includegraphics[width=\linewidth]{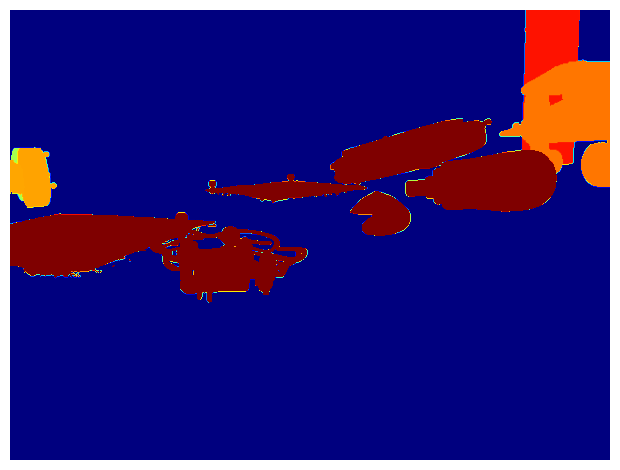}
            &\includegraphics[width=\linewidth]{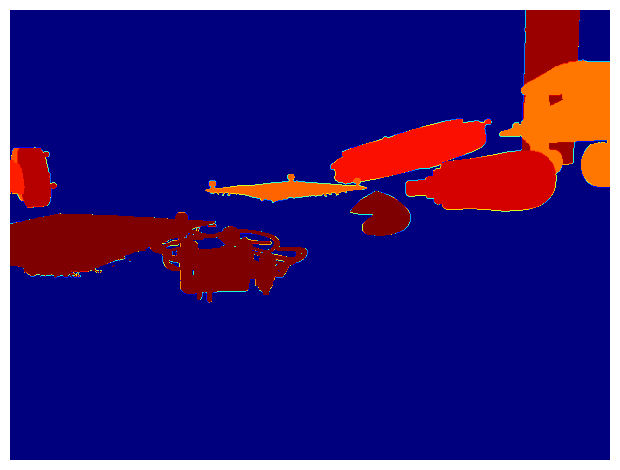}
            &\includegraphics[width=\linewidth]{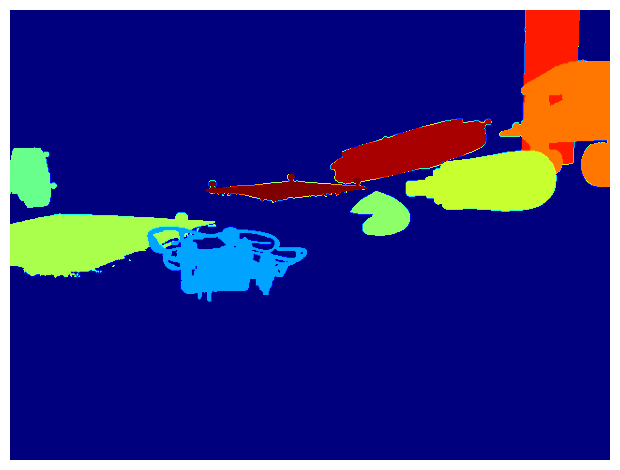}
            &\includegraphics[width=\linewidth]{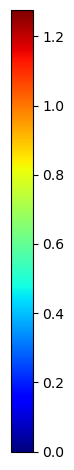}
		\\

         (a) & (b) & (c) & (d) & (e)
        \\
	\end{tabular}
	}
    \vspace{-1ex}
    \caption{Qualitative results of depth estimation on the EVIMO2 dataset, with framewise results on four exemplary video sequences: (a) Events and segmentation mask. (b) Ground truth. (c) E2Depth~\cite{hidalgo2020learning}. (d) EMVS~\cite{rebecq2018emvs}. (e) Ours.
    \label{images:depth_comp2}
    }
\end{figure}

\def\figWidth2{0.19\linewidth}
\def\smallfigWidth{0.19\linewidth}

\noindent {\bf Object-wise distance estimation.} All monocular depth estimation methods, including ours, rely solely on visual information from events without using frame-based data.
Unlike traditional multi-view stereo methods requiring known camera viewpoints, EMVS~\cite{rebecq2018emvs} estimates semi-dense 3D structures from event cameras using known motion trajectories.Traditional methods require absolute positional information, such as trajectories or camera poses. In contrast, our approach only needs relative camera motion, specifically angular velocity, which is readily available from on-board IMU sensors, unlike absolute camera pose that needs external systems like Motion Capture. Hidalgo-Carri\'{o} et. al use only events to learn a dense depth map. Unlike earlier learning-based methods, they propose a recurrent network architecture to maintain temporal consistency.

Qualitative results for object-wise relative depth estimation are presented in \cref{images:depth_comp1}.
We present relative depth estimation results at two distinct timestamps: 11.5s and 13.5s, respectively. The GT relative depth is shown and compared to three different approaches that perform monocular depth estimation from an event stream: E2Depth~\cite{hidalgo2020learning}, EMVS~\cite{rebecq2018emvs} and Ours. While each method yields depth estimates in various formats - dense depth, sparse depth, and object-wise depth - we unify results into a common format for consistent evaluation.
In addition, more comparisons are shown in \cref{images:depth_comp2}, where our method (column d) reports the closest results to GT (column e) while E2depth and EMVS deviate a lot (column b-c).

\Cref{table:main-results} shows the quantitative comparison of accuracy on all test sequences of the structure-from-motion (sfm) and the structure-from-motion split in low light conditions (sfm low light) of EVIMO2. On \textit{smf}, our approach improves RMSE (linear) by over 16\%. On \textit{sfm low light}, it achieves a 5\% improvement. In terms of RMSE (log), E2Depth outperforms all other methods. The performance gain on the \textit{smf} split is less pronounced in low light conditions. We identified two reasons for this lower performance. First, low light conditions cause significantly higher event sparsity. Second, the ratio of hot pixels to informative events deteriorates. Hot pixels are sensor failures that consistently ``fire'' regardless of camera or scene motion.
Compared to EMVS and E2Depth, our approach, which relies on \textit{active event alignment}, can be affected by pixel failures in the dynamic vision sensor. Although hot pixels, being locally stable, can hinder event alignment, our remarkable 16\% performance gain highlights the efficacy of combining dynamic vision systems like event cameras with active vision approaches.

\textbf{Object-wise event alignment} utilises a modified version of the original Spatio-Temporal Poisson Point Process for distance estimation. 
While the original version performs event alignment given a fixed number of events - typically 30K events~\cite{gallego2018unifying,gallego2018unifying, Nunes20eccv, Gu_2021}, we perform event alignment of all events within a fixed time interval $\Delta T$.
~\Cref{table:Angular_velocity_estimation} shows, that defining the Poisson Point Process for a fixed time interval not only improves performance, also it is consistent with the original definition of the Poisson Point Process~\cite{streit2010poisson}.

\begin{table}[hb]
\raggedright
\small
\begin{adjustbox}{max width=\linewidth}
\setlength{\tabcolsep}{3pt}
\begin{tabular}{p{2ex}lcccccc}
\toprule 

& \bf{Method} & $e_{wx}$ & $e_{wy}$ & $e_{wz}$ & $\sigma_{ew}$ & RMS & RMS\% \\
\midrule 
\multirow{5}{*}{\rotatebox{90}{\makecell{\emph{boxes}}}} 
& CMax \cite{gallego2018unifying} & 7.38 & 6.66 & 6.03 & 9.04 & 9.08 & 0.66\\
& AEMin \cite{Nunes20eccv} & 6.75 & 5.19 & 5.78 & 7.77 & 7.81 & 0.56\\
& EMin \cite{Nunes20eccv} & 6.55 & 4.40 & 5.00 & 7.00 & 7.06 & 0.51 \\ \addlinespace[0.02cm]
& Ours ($Ne = 30K$) & 6.72 & 3.93 & 4.55 & 6.64 & 6.73 & 0.49\\
& Ours ($\Delta T = 0.015$) & \textbf{5.68} & \textbf{3.81} & \textbf{3.92} & \textbf{6.32} & \textbf{6.34} & \textbf{0.46} \\
\midrule 
\multirow{5}{*}{\rotatebox{90}{\makecell{\emph{poster}}}} 
& CMax \cite{gallego2018unifying} & 13.45 & 9.87 & 5.56 & 13.39 & 13.45 & 0.74\\
& AEMin \cite{Nunes20eccv} & 12.57 & 7.89 & 5.63 & 12.35 & 12.36 & 0.68\\
& EMin \cite{Nunes20eccv} & 11.83 & 7.31 & 4.37 & 10.85 & 10.86 & 0.60 \\ \addlinespace[0.02cm]
& Ours ($Ne = 30K$) & 11.78 & 6.33 & 3.67 & 10.30 & 10.37 & 0.57\\ 
& Ours ($\Delta T = 0.015$) & \textbf{9.37} & \textbf{5.77} & \textbf{3.49} & \textbf{9.15} & \textbf{9.21} & \textbf{0.51}\\
\midrule
\multirow{5}{*}{\rotatebox{90}{\makecell{\emph{dynamic}}}}
& CMax \cite{gallego2018unifying} & 4.93 & 4.82 & 4.95 & 7.11 & 7.13 & 0.71\\
& AEMin\cite{Nunes20eccv} & 5.02 & 3.88 & 4.55 & 6.16 & 6.19 & 0.62\\
& EMin \cite{Nunes20eccv} & 4.78 & 3.72 & 3.73 & 5.33 & 5.39 & 0.54\\
& Ours ($Ne = 30K$) & 4.42 & 3.61 & \textbf{3.49} & \textbf{5.15} & \textbf{5.19} & \textbf{0.52}\\
& Ours ($\Delta T = 0.015$) & \textbf{4.29} & \textbf{3.60} & 3.97 & 5.27 & 5.33 & 0.53\\
\midrule
\multirow{5}{*}{\rotatebox{90}{\makecell{\emph{shapes}}}}
& CMax \cite{gallego2018unifying} & 31.19 & 26.83 & 38.98 & 55.86 & 55.87 & 3.94 \\
& AEMin \cite{Nunes20eccv} & 22.22 & 18.78 & 35.41 & 55.43 & 55.44 & 3.91 \\
& EMin \cite{Nunes20eccv} & 21.22 & 15.87 & 25.57 & 42.22 & 42.22 & 2.98 \\
& Ours ($Ne = 30K$) & 20.73 & 13.95 & 17.69 & 25.88 & 25.89 & 1.83 \\
& Ours ($\Delta T = 0.015$) & \textbf{10.32} & \textbf{5.61} & \textbf{4.68} & \textbf{10.15} & \textbf{10.16} & \textbf{0.69}\\
\bottomrule
\end{tabular}
\end{adjustbox}
\vspace{0.5ex}
\caption{
Comparison of angular velocity accuracy on the rotation sequences from DAVIS 240C~\cite{Mueggler17ijrr}.}
\label{table:Angular_velocity_estimation}
\end{table}

\subsection{Ablation Study}

In \cref{sec:results} we discussed our results on object-wise relative distance estimation. These results are based on two key assumptions: (1) the presence of fronto-parallel planar object regions, and (2) zero translational motion along the depth axis (refer to~\cref{subsubsec:distance}). Here we ask, \textit{How sensitive is our algorithm to the quality of predefined object regions?} and \textit{How does z-motion affect the accuracy of relative depth estimates?}
To address these questions, we first present qualitative results of relative depth estimation using segmentation masks that are entirely object-independent. Then, we examine our algorithm's performance, with respect to its sensitivity to camera motion, with a particular focus on z-motion.

\begin{figure}
  \centering
  \begin{subfigure}{0.19\linewidth}
  \includegraphics[height=1.2cm, width=1.6cm]{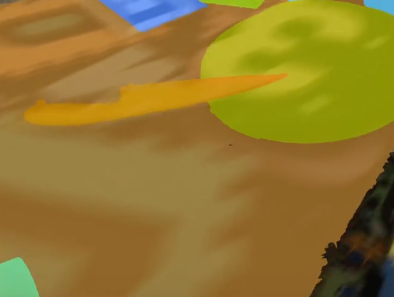}\\
  \includegraphics[height=1.2cm, width=1.6cm]{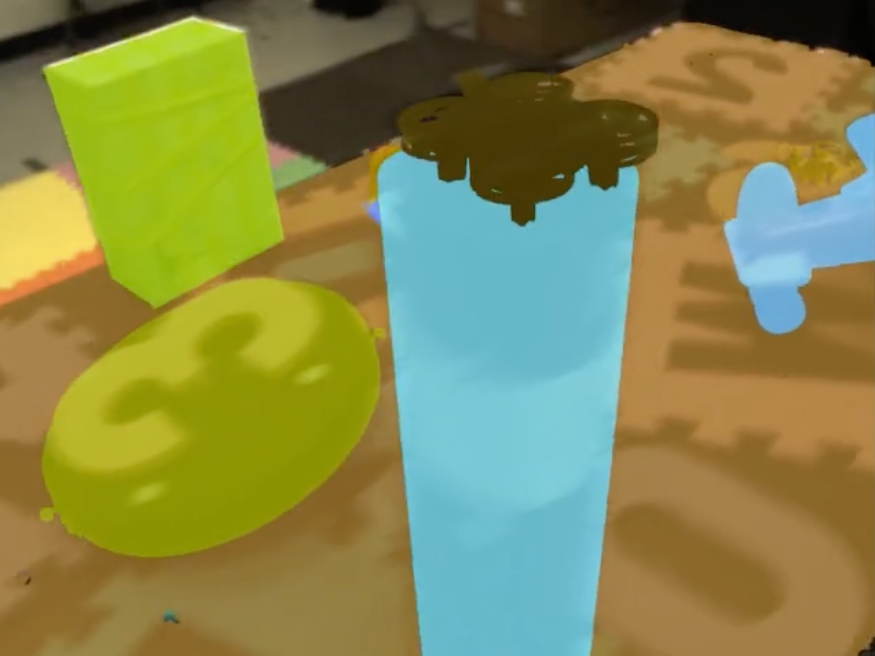}\\
  \includegraphics[height=1.2cm, width=1.6cm]{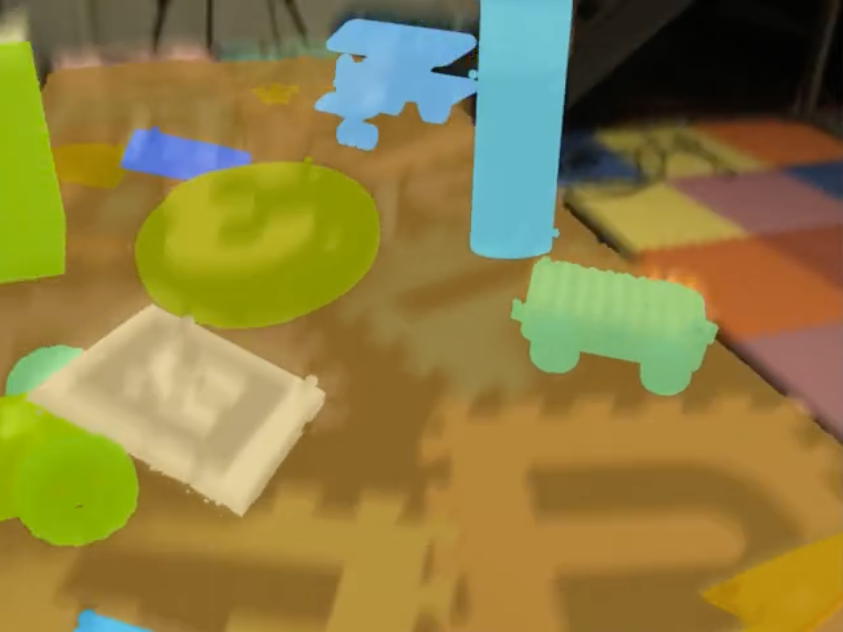}
  \subcaption{}
  \end{subfigure}
  \begin{subfigure}{0.19\linewidth}
  \includegraphics[height=1.2cm, width=1.6cm]{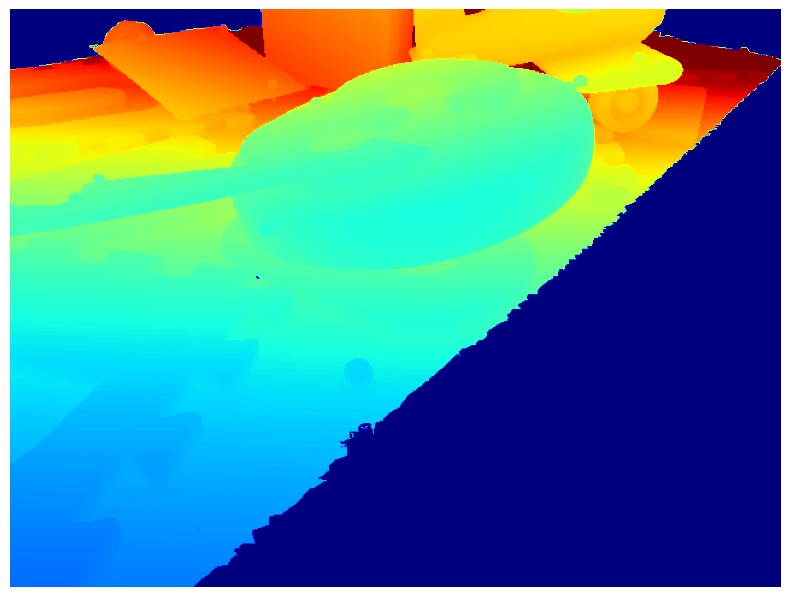}\\
  \includegraphics[height=1.2cm, width=1.6cm]{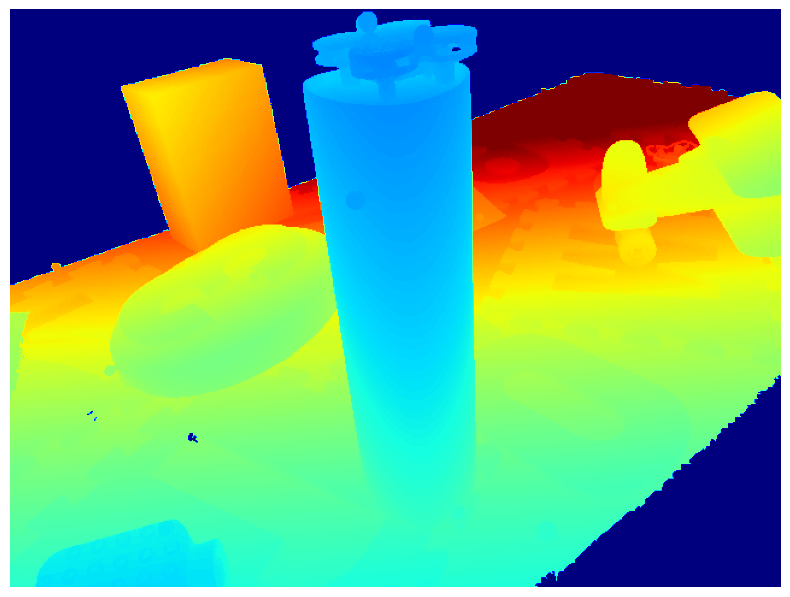}\\
  \includegraphics[height=1.2cm, width=1.6cm]{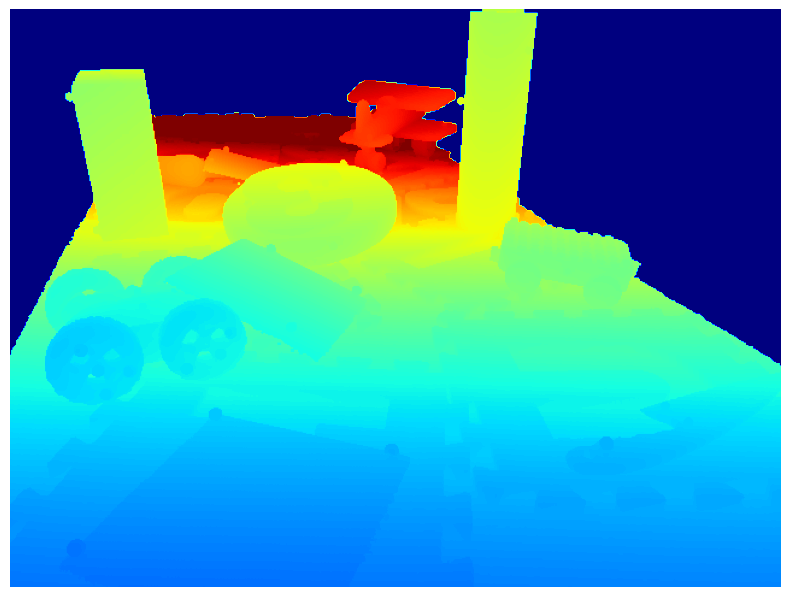}
  \subcaption{}
  \end{subfigure}
  \begin{subfigure}{0.19\linewidth}
  \includegraphics[height=1.2cm, width=1.6cm]{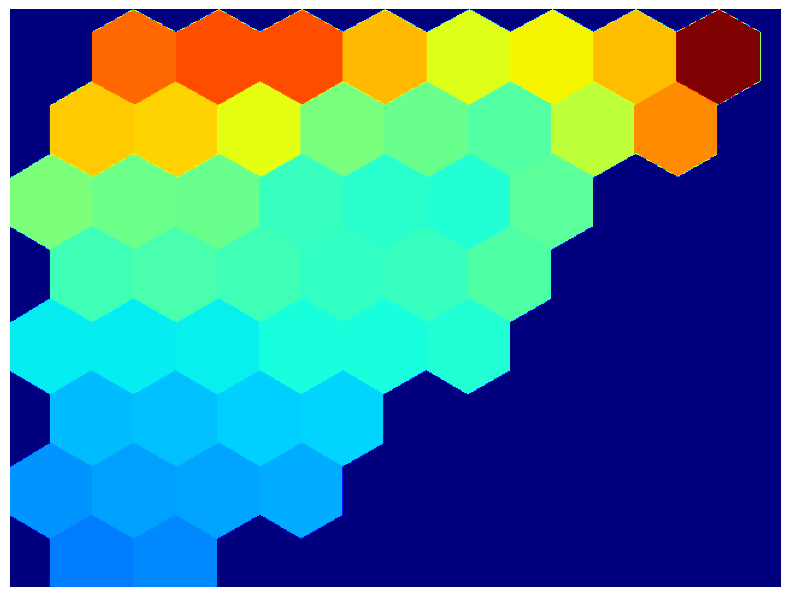}\\
  \includegraphics[height=1.2cm, width=1.6cm]{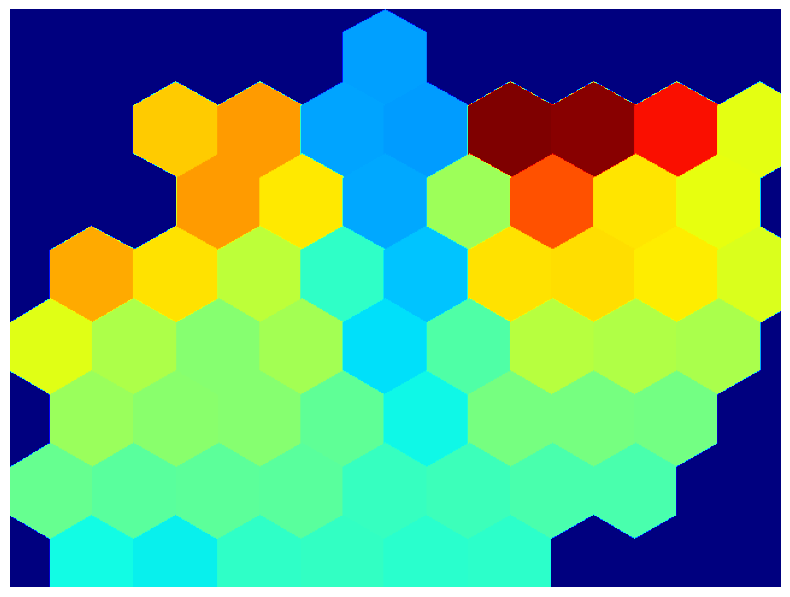}\\
  \includegraphics[height=1.2cm, width=1.6cm]{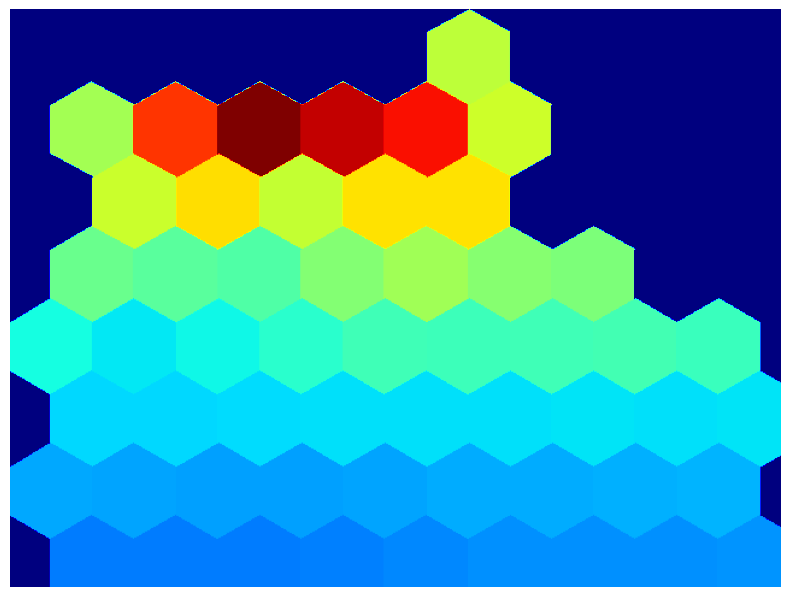}
  \subcaption{}
  \end{subfigure}
  \begin{subfigure}{0.19\linewidth}
  \includegraphics[height=1.2cm, width=1.6cm]{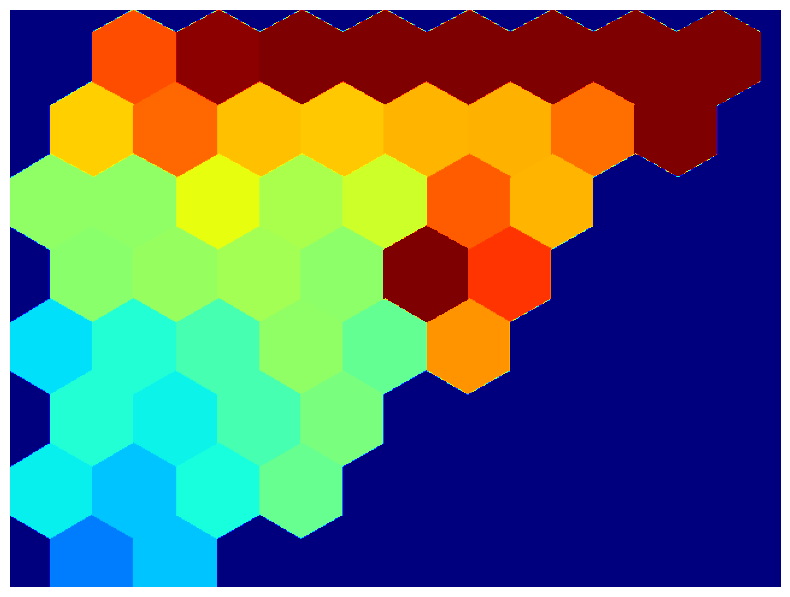}\\
  \includegraphics[height=1.2cm, width=1.6cm]{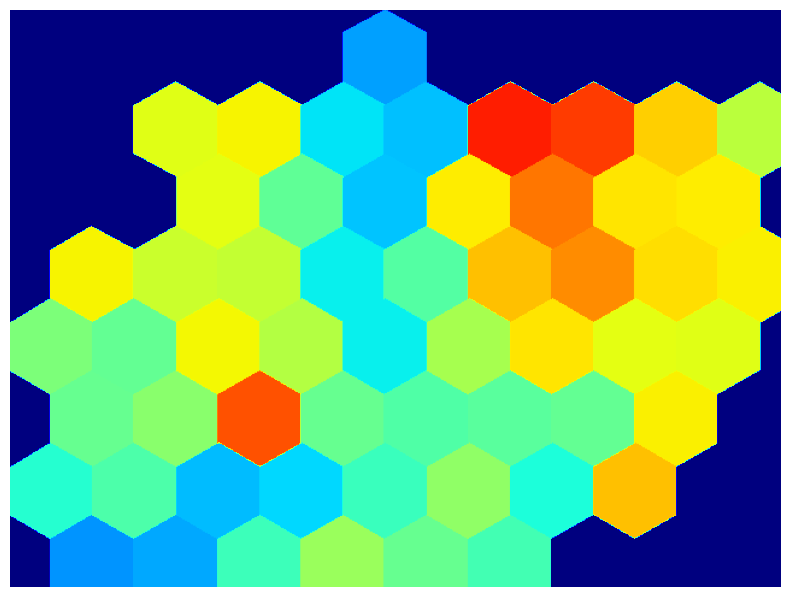}\\
  \includegraphics[height=1.2cm, width=1.6cm]{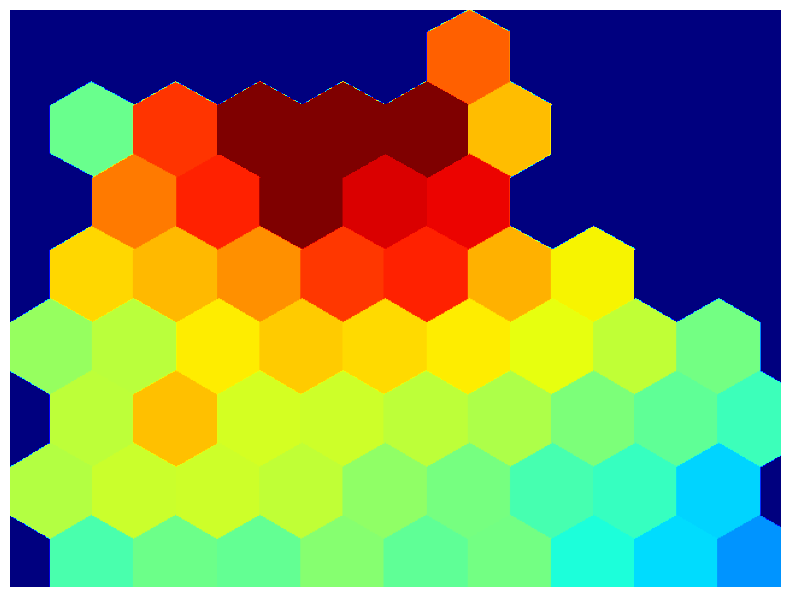}
  \subcaption{}
  \end{subfigure}
  \begin{subfigure}{0.19\linewidth}
  \includegraphics[height=1.2cm, width=1.6cm]{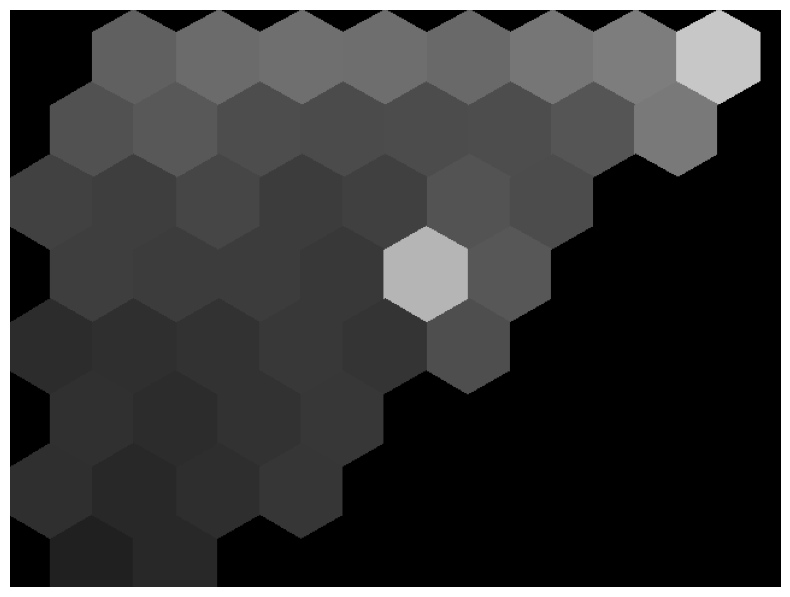}\\
  \includegraphics[height=1.2cm, width=1.6cm]{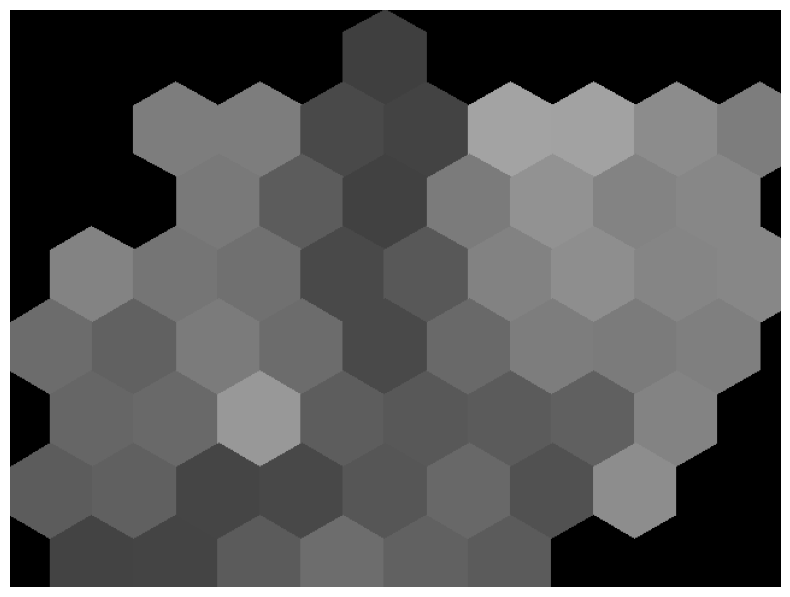}\\
  \includegraphics[height=1.2cm, width=1.6cm]{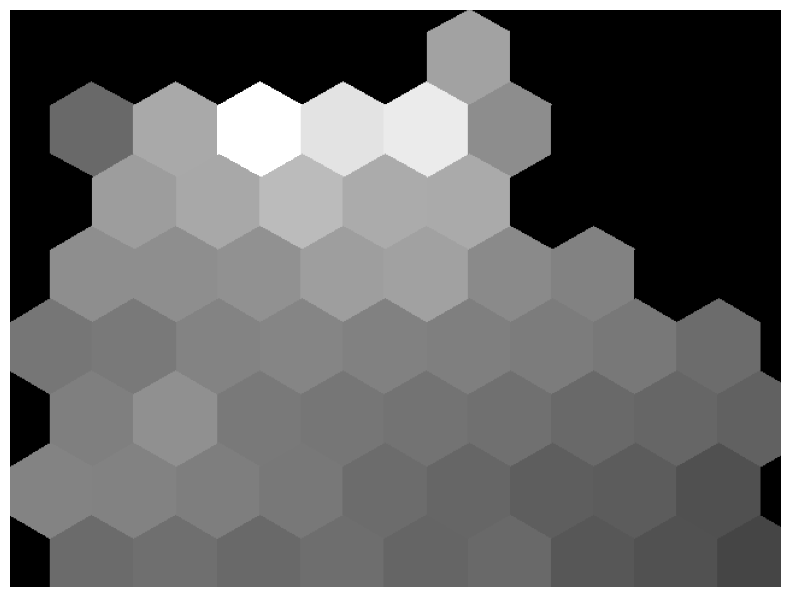}
  \subcaption{}
  \label{images-sub:confidence}
  \end{subfigure}
  \caption{Qualitative evaluation of region-wise distance estimation w/o object masks. We use a honeycomb grid to define pixel regions for depth estimation. Relative distances and confidence maps are shown in grayscale (white = low confidence, black = high confidence).
  (a) Events and segmentation masks.
  (b) Original ground truth.
  (c) Ground truth using honeycomb regions.
  (d) Our method using honeycomb regions.
  (e) $\pm 3\sigma$ confidence interval.
  }
  \label{images:honeycomb}
\end{figure}

\noindent{\bf Object regions.}
\cref{images:honeycomb} qualitatively shows region-wise depth estimation results of our algorithm. To avoid relying on informative object masks, that may provide strong priors in terms of depth consistency within a particular region, we employed a honeycomb grid to define pixel regions independent of any object or scene information. We show that even without properly defined image regions we can determine the relative depth reasonably well. To the best of our knowledge our algorithm for the first time models the estimate's confidence plus tracks the confidence over time. The $\pm 3\sigma$ confidence interval is shown in ~\cref{images-sub:confidence}. It nicely captures deviating estimates from the GT. The resolution of the acquired depth estimates is undoubtedly influenced by the grid size of the masks. However, prior work emphasizes that region-wise estimates are crucial for distance estimation, while precise object boundaries are not essential~\cite{ballard1990animate, mishra2009active}.

\begin{figure}[t]
    \centering
    \includegraphics[width=0.2\textwidth]{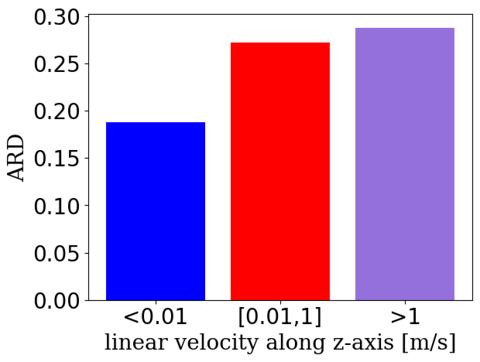}
    \caption{Evaluation of distance estimates based on the camera's z-motion: low, normal, and high speed.}
  \label{fig:assumption1}
\end{figure}

\noindent{\bf Zero translational motion along depth axis.}
The assumption of zero linear velocity along the z-axis (no forward/backward camera motion) allows the approximation of a camera translation with a rotation. 
However, the EVIMO2 dataset includes event sequences with arbitrary camera movement, meaning our assumption of $W=0$ holds only in very few cases. 
\cref{fig:assumption1} evaluates distance estimation via active alignment dependent upon the amount of linear velocity along the z-axis. As expected, error decreases with lower z-axis velocity, aligning better with our assumption.

\begin{figure}[H]
  \centering
  \begin{subfigure}{.3\linewidth}
  \centering
  \rotatebox{180}{\includegraphics[height=1.8cm]{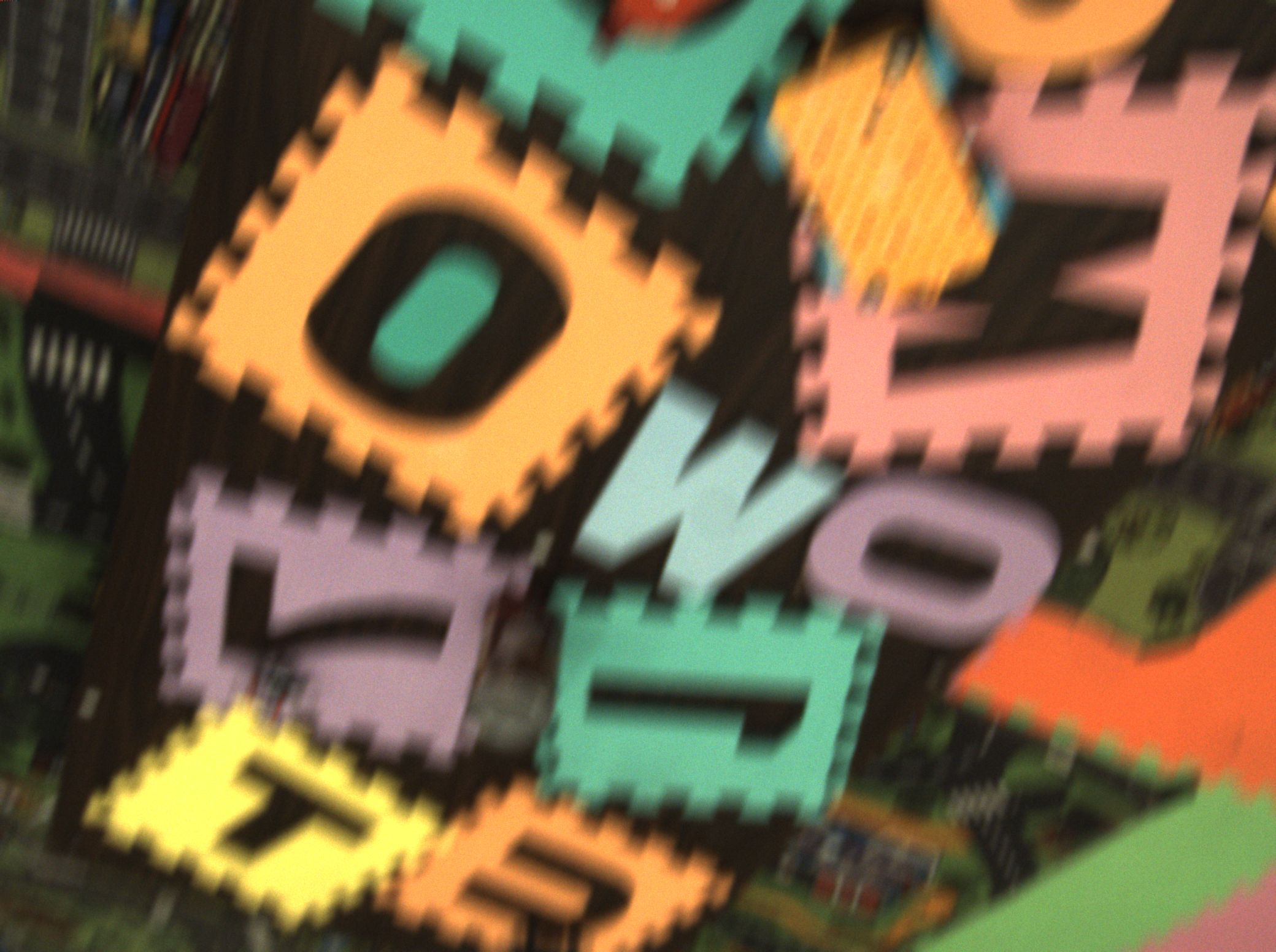}}
  \caption{RGB frame \textcolor{white}{bla bla}}
  \end{subfigure}
  \begin{subfigure}{.3\linewidth}
  \centering
  \includegraphics[height=1.8cm]{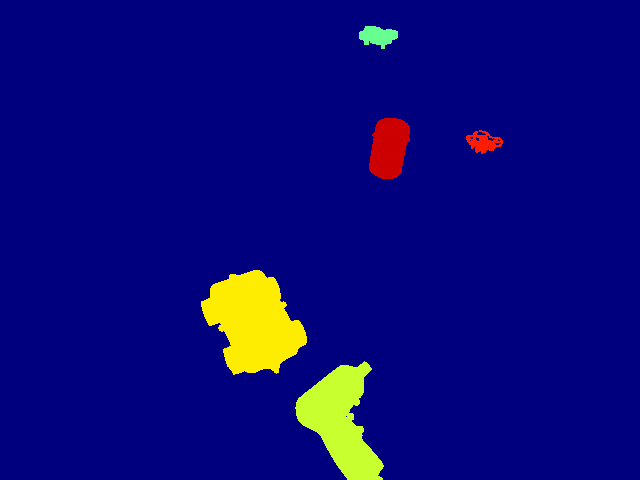}
  \caption{Estimated relative depth $d_t$}
  \end{subfigure}
  \begin{subfigure}{.3\linewidth}
  \centering
  \includegraphics[height=1.8cm]{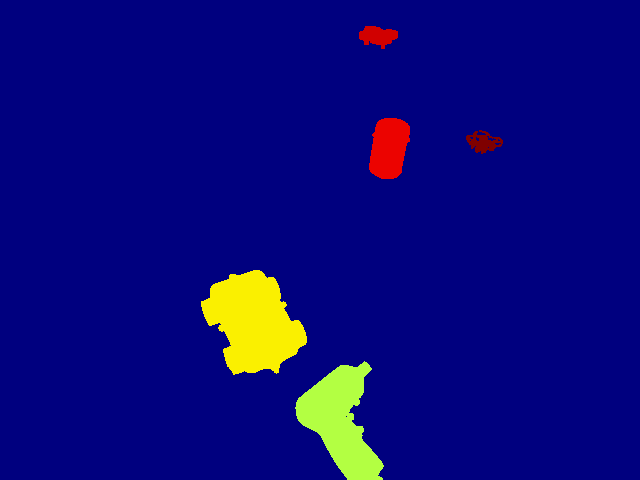}
  \caption{Ground truth\\ relative depth $d_t$}
  \end{subfigure}
  \caption{Failure case: Relative distance estimation fails for moving objects, as illustrated by the flying drone.}
  \label{images:failure_case}
\end{figure}

\paragraph{Failure case analysis.} Relative distance is estimated between two static objects. If one object is moving, the distance cannot be accurately inferred. Interestingly, relative depth estimates between neighboring static objects remain unaffected. This suggests potential new research directions, such as detecting object motion by analyzing discontinuities in relative distances between multiple objects. Specifically, given three objects, we can compute their relative distances. If these distances remain constant, none of the object is moving. Conversely, if the relative distances change, at least one of the objects is in motion. \cref{images:failure_case} shows that, in the presence of object motion, the relative depth estimates of neighboring static objects remain unaffected, while the estimated relative distance of the moving object diverges from the ground truth.

\section{Conclusion}
Inspired by visual ecology, we propose the first event-based approach for relative distance estimation that combines dynamic vision sensors with a behavioral strategy to infer relative distances between objects. Firstly, we introduce a novel optimization pipeline that estimates a rotational motion aimed at achieving \textit{object-wise event alignment}. This rotation does not recover the actual camera motion but is a virtual adjustment designed to align events. Secondly, object-wise \textit{relative distance} is determined by comparing the corresponding rotational flow vectors. 

Compared to frame-based cameras event cameras capture visual information efficiently by reducing redundant data. However, processing methods are still developing. Active alignment, which computes new motions to extract visual data like depth, bear a high potential for novel, efficient vision algorithms. Our approach reduces computational load by using behavioral strategies, such as gaze stabilization, to streamline sensory input processing.

%
%
{\small
\bibliographystyle{ieee_fullname}
\bibliography{egbib}

\begin{thebibliography}{10}\itemsep=-1pt

\bibitem{aloimonos1988active}
John Aloimonos, Isaac Weiss, and Amit Bandyopadhyay.
\newblock Active vision.
\newblock {\em IJCV}, 1:333--356, 1988.

\bibitem{ballard1990animate}
Dana~H Ballard.
\newblock Animate vision uses object-centered reference frames.
\newblock In {\em Advanced neural computers}, pages 229--236. Elsevier, 1990.

\bibitem{barlow1999finding}
H.B. Barlow, T.P. Kaushal, and G.J. Mitchison.
\newblock {Finding Minimum Entropy Codes}.
\newblock {\em Neural Computation}, 1(3):412--423, 09 1989.

\bibitem{barlow1989unsupervised}
Horace~B Barlow.
\newblock Unsupervised learning.
\newblock {\em Neural computation}, 1(3):295--311, 1989.

\bibitem{battaje2022one}
Aravind Battaje and Oliver Brock.
\newblock One object at a time: Accurate and robust structure from motion for robots.
\newblock In {\em IEEE/RSJ Int. Conf. Intell. Robot. Syst. (IROS)}, pages 3598--3603. IEEE, 2022.

\bibitem{Bideau2024}
Pia Bideau, Erik Learned-Miller, Cordelia Schmid, and Karteek Alahari.
\newblock The right spin: Learning object motion from rotation-compensated flow fields.
\newblock {\em International Journal of Computer Vision}, 132(1):40--55, Jan 2024.

\bibitem{EVIMO2}
L. {Burner}, A. {Mitrokhin}, C. {Ye}, C. {Ferm\"uller}, Y. {Aloimonos}, and T. {Delbruck}.
\newblock Evimo2: An event camera dataset for motion segmentation, optical flow, structure from motion, and visual inertial odometry in indoor scenes with monocular or stereo algorithms.
\newblock In {\em arXiv e-prints}, May 2022.

\bibitem{burner2023ttcdist}
Levi Burner, Nitin~J Sanket, Cornelia Ferm{\"u}ller, and Yiannis Aloimonos.
\newblock Ttcdist: Fast distance estimation from an active monocular camera using time-to-contact.
\newblock In {\em {IEEE} Int. Conf. Robot. Autom. (ICRA)}, pages 4909--4915. IEEE, 2023.

\bibitem{Cabriel2023}
Cl{\'e}ment Cabriel, Tual Monfort, Christian~G. Specht, and Ignacio Izeddin.
\newblock Event-based vision sensor for fast and dense single-molecule localization microscopy.
\newblock {\em Nature Photonics}, 17(12):1105--1113, Dec 2023.

\bibitem{DSI}
R.T. Collins.
\newblock A space-sweep approach to true multi-image matching.
\newblock In {\em CVPR}, pages 358--363, 1996.

\bibitem{2014depth}
David Eigen, Christian Puhrsch, and Rob Fergus.
\newblock Depth map prediction from a single image using a multi-scale deep network.
\newblock {\em NeurIPS}, 27, 2014.

\bibitem{fermuller1992tracking}
Cornelia Ferm{\"u}ller and Yiannis Aloimonos.
\newblock Tracking facilitates 3-d motion estimation.
\newblock {\em Biological Cybernetics}, 67(3):259--268, 1992.

\bibitem{fermuller1993role}
Cornelia Ferm{\"u}ller and Yiannis Aloimonos.
\newblock The role of fixation in visual motion analysis.
\newblock {\em International Journal of Computer Vision}, 11(2):165--186, 1993.

\bibitem{field1994goal}
David~J Field.
\newblock What is the goal of sensory coding?
\newblock {\em Neural computation}, 6(4):559--601, 1994.

\bibitem{gallego2019focus}
Guillermo Gallego, Mathias Gehrig, and Davide Scaramuzza.
\newblock Focus is all you need: Loss functions for event-based vision.
\newblock In {\em CVPR}, pages 12280--12289, 2019.

\bibitem{gallego2018unifying}
Guillermo Gallego, Henri Rebecq, and Davide Scaramuzza.
\newblock A unifying contrast maximization framework for event cameras, with applications to motion, depth, and optical flow estimation.
\newblock In {\em CVPR}, pages 3867--3876, 2018.

\bibitem{gallego2017accurate}
Guillermo Gallego and Davide Scaramuzza.
\newblock Accurate angular velocity estimation with an event camera.
\newblock {\em IEEE Robotics and Automation Letters}, 2(2):632--639, 2017.

\bibitem{Gehrig2024}
Daniel Gehrig and Davide Scaramuzza.
\newblock Low-latency automotive vision with event cameras.
\newblock {\em Nature}, 629(8014):1034--1040, May 2024.

\bibitem{Ghosh_2022}
Suman Ghosh and Guillermo Gallego.
\newblock Multi-event-camera depth estimation and outlier rejection by refocused events fusion.
\newblock {\em Advanced Intelligent Systems}, page 2200221, sep 2022.

\bibitem{godard2019digging}
Cl{\'e}ment Godard, Oisin Mac~Aodha, Michael Firman, and Gabriel~J Brostow.
\newblock Digging into self-supervised monocular depth estimation.
\newblock In {\em ICCV}, pages 3828--3838, 2019.

\bibitem{Gu_2021}
Cheng Gu, Erik Learned-Miller, Daniel Sheldon, Guillermo Gallego, and Pia Bideau.
\newblock The spatio-temporal poisson point process: A simple model for the alignment of event camera data.
\newblock In {\em ICCV}. {IEEE}, oct 2021.

\bibitem{hassan2024efficient}
Eman Hassan, Zhuowen Zou, Hanning Chen, Mohsen Imani, Yahya Zweiri, Hani Saleh, and Baker Mohammad.
\newblock Efficient event-based robotic grasping perception using hyperdimensional computing.
\newblock {\em Internet of Things}, 26:101207, 2024.

\bibitem{hidalgo2020learning}
Javier Hidalgo-Carri{\'o}, Daniel Gehrig, and Davide Scaramuzza.
\newblock Learning monocular dense depth from events.
\newblock In {\em Int. Conf. 3D Vision (3DV)}, pages 534--542. IEEE, 2020.

\bibitem{Horn1981}
Berthold K.~P. Horn and Brian~G. Schunck.
\newblock Determining optical flow.
\newblock {\em Artificial Intelligence}, 17(1-3):185--203, 1981.

\bibitem{hwang2023ev}
Inwoo Hwang, Junho Kim, and Young~Min Kim.
\newblock Ev-nerf: Event based neural radiance field.
\newblock In {\em Proceedings of the IEEE/CVF Winter Conference on Applications of Computer Vision}, pages 837--847, 2023.

\bibitem{jiang2018self}
Huaizu Jiang, Gustav Larsson, Michael Maire~Greg Shakhnarovich, and Erik Learned-Miller.
\newblock Self-supervised relative depth learning for urban scene understanding.
\newblock In {\em ECCV}, pages 19--35, 2018.

\bibitem{HOTS}
Xavier Lagorce, Garrick Orchard, Francesco Galluppi, Bertram~E. Shi, and Ryad~B. Benosman.
\newblock Hots: A hierarchy of event-based time-surfaces for pattern recognition.
\newblock {\em IEEE Transactions on Pattern Analysis and Machine Intelligence}, 39(7):1346--1359, 2017.

\bibitem{land2012animal}
Michael~F Land and Dan-Eric Nilsson.
\newblock {\em Animal eyes}.
\newblock OUP Oxford, 2012.

\bibitem{mishra2009active}
Ajay Mishra, Yiannis Aloimonos, and Cornelia Fermuller.
\newblock Active segmentation for robotics.
\newblock In {\em IEEE/RSJ Int. Conf. Intell. Robot. Syst. (IROS)}, pages 3133--3139. IEEE, 2009.

\bibitem{mithokinIROS18}
Anton Mitrokhin, Cornelia Fermüller, Chethan Parameshwara, and Yiannis Aloimonos.
\newblock Event-based moving object detection and tracking.
\newblock In {\em IEEE/RSJ Int. Conf. Intell. Robot. Syst. (IROS)}, pages 1--9, 2018.

\bibitem{Mueggler17ijrr}
Elias Mueggler, Henri Rebecq, Guillermo Gallego, Tobi Delbruck, and Davide Scaramuzza.
\newblock The event-camera dataset and simulator: Event-based data for pose estimation, visual odometry, and {SLAM}.
\newblock {\em Int. J. Robot. Research}, 36(2):142--149, 2017.

\bibitem{mueggler2017event}
Elias Mueggler, Henri Rebecq, Guillermo Gallego, Tobi Delbruck, and Davide Scaramuzza.
\newblock The event-camera dataset and simulator: Event-based data for pose estimation, visual odometry, and slam.
\newblock {\em The International Journal of Robotics Research}, 36(2):142--149, 2017.

\bibitem{Nunes20eccv}
Urbano~Miguel Nunes and Yiannis Demiris.
\newblock Entropy minimisation framework for event-based vision model estimation.
\newblock In {\em ECCV}, 2020.

\bibitem{rebecq2018emvs}
Henri Rebecq, Guillermo Gallego, Elias Mueggler, and Davide Scaramuzza.
\newblock Emvs: Event-based multi-view stereo—3d reconstruction with an event camera in real-time.
\newblock {\em IJCV}, 126(12):1394--1414, 2018.

\bibitem{roth2023objects}
Nicolas Roth, Martin Rolfs, Olaf Hellwich, and Klaus Obermayer.
\newblock Objects guide human gaze behavior in dynamic real-world scenes.
\newblock {\em bioRxiv}, pages 2023--03, 2023.

\bibitem{rudnev2023eventnerf}
Viktor Rudnev, Mohamed Elgharib, Christian Theobalt, and Vladislav Golyanik.
\newblock Eventnerf: Neural radiance fields from a single colour event camera.
\newblock In {\em CVPR}, pages 4992--5002, 2023.

\bibitem{shi2023even}
Peilun Shi, Jiachuan Peng, Jianing Qiu, Xinwei Ju, Frank Po~Wen Lo, and Benny Lo.
\newblock Even: An event-based framework for monocular depth estimation at adverse night conditions.
\newblock {\em arXiv preprint arXiv:2302.03860}, 2023.

\bibitem{stoffregen2019event}
Timo Stoffregen, Guillermo Gallego, Tom Drummond, Lindsay Kleeman, and Davide Scaramuzza.
\newblock Event-based motion segmentation by motion compensation.
\newblock In {\em ICCV}, pages 7244--7253, 2019.

\bibitem{streit2010poisson}
Roy~L Streit and Roy~L Streit.
\newblock {\em The poisson point process}.
\newblock Springer, 2010.

\bibitem{thrun2002probabilistic}
Sebastian Thrun.
\newblock Probabilistic robotics.
\newblock {\em Communications of the ACM}, 45(3):52--57, 2002.

\bibitem{zhou2021event_stereoVO}
Yi Zhou, Guillermo Gallego, and Shaojie Shen.
\newblock Event-based stereo visual odometry.
\newblock {\em IEEE Transactions on Robotics}, 37(5):1433--1450, 2021.

\bibitem{zhu2019unsupervised}
Alex~Zihao Zhu, Liangzhe Yuan, Kenneth Chaney, and Kostas Daniilidis.
\newblock Unsupervised event-based learning of optical flow, depth, and egomotion.
\newblock In {\em CVPR}, pages 989--997, 2019.

\bibitem{zhu2023self}
Junyu Zhu, Lina Liu, Bofeng Jiang, Feng Wen, Hongbo Zhang, Wanlong Li, and Yong Liu.
\newblock Self-supervised event-based monocular depth estimation using cross-modal consistency.
\newblock In {\em IEEE/RSJ Int. Conf. Intell. Robot. Syst. (IROS)}, pages 7704--7710. IEEE, 2023.

\end{thebibliography}
}

\end{document}